\documentclass{article}

\usepackage[table]{xcolor}

\usepackage[numbers]{natbib}

\usepackage[preprint]{neurips_2025}

\usepackage{wrapfig}

\usepackage{listings}
\usepackage{paralist}
\usepackage{stix}
\usepackage{natbib}

\usepackage{hyperref}       % hyperlinks

\usepackage[inline]{enumitem}
\usepackage{multirow}

\usepackage{arydshln}

\usepackage{booktabs}
\usepackage{comment}
\usepackage{nicematrix}

\usepackage{graphicx,mathtools,kantlipsum}

\usepackage{subcaption}

\usepackage{pifont}% http://ctan.org/pkg/pifont
\newcommand{\cmark}{\ding{51}}%
\newcommand{\xmark}{\ding{55}}%

\definecolor{GREENBACK}{HTML}{F1FFF3}
\definecolor{NEGATIVE}{HTML}{4a4e69}
\definecolor{POSITIVE}{HTML}{32CD32}
\definecolor{GREENBACK}{HTML}{95d5b2}
\definecolor{dkgreen}{rgb}{0,0.6,0}
\definecolor{gray}{rgb}{0.5,0.5,0.5}
\definecolor{mauve}{rgb}{0.58,0,0.82}

\usepackage[utf8]{inputenc} % allow utf-8 input
\usepackage[T1]{fontenc}    % use 8-bit T1 fonts

\usepackage{url}            % simple URL typesetting
\usepackage{booktabs}       % professional-quality tables
\usepackage{amsfonts}       % blackboard math symbols
\usepackage{nicefrac}       % compact symbols for 1/2, etc.
\usepackage{microtype}      % microtypography

\usepackage[nameinlink]{cleveref}
\usepackage{arydshln}
\usepackage{tikz}
\usepackage{etoolbox}

\title{ReCo: Reminder Composition Mitigates Hallucinations in Vision-Language Models}

\author{%
  Sotirios Panagiotis Chytas \thanks{chytas@wisc.edu, \url{https://github.com/SPChytas/ReCo}} \\
  University of Wisconsin-Madison\\
  \And
  Miso Choi \\
  Korea University\\
  \AND
  Hyunwoo J. Kim \\
  KAIST\\
  \And
  Vikas Singh \\
  University of Wisconsin-Madison\\
}

\begin{document}

\maketitle

\begin{abstract}
Vision Language Models (VLMs) show impressive 
capabilities in integrating and reasoning with both 
visual and language data.
But these models make mistakes. 
A common finding -- similar to LLMs -- is their tendency to hallucinate, i.e., generate plausible sounding 
text which is not grounded in the visual input, or at worst, 
is contradictory.
A growing consensus attributes this behavior to 
an over-reliance on language -- especially as the generation progresses, the model suffers from 
a ``fading memory effect'' with respect to the provided visual input.
We study mechanisms by which this behavior can 
be controlled. Specifically, 
using ideas from geometric algebra and relational compositions, we propose the addition of a small, trainable module 
(named ReCo) on top of any VLM -- no other modification is needed. 
We show that such a lightweight module is able to mitigate the fading memory effect on three of the most widely used VLMs (InstructBLIP, LlaVA, MiniGPT4), where we see 
performance improvements on multiple benchmarks. 
Additionally, we show that our module can be combined with many of the other approaches for reducing hallucination where we achieve improved results for each one. 
\end{abstract}

\section{Introduction}

\begin{wrapfigure}{r}{0.45\textwidth}
    \centering
    \vspace{-15pt}
    \includegraphics[width=\linewidth]{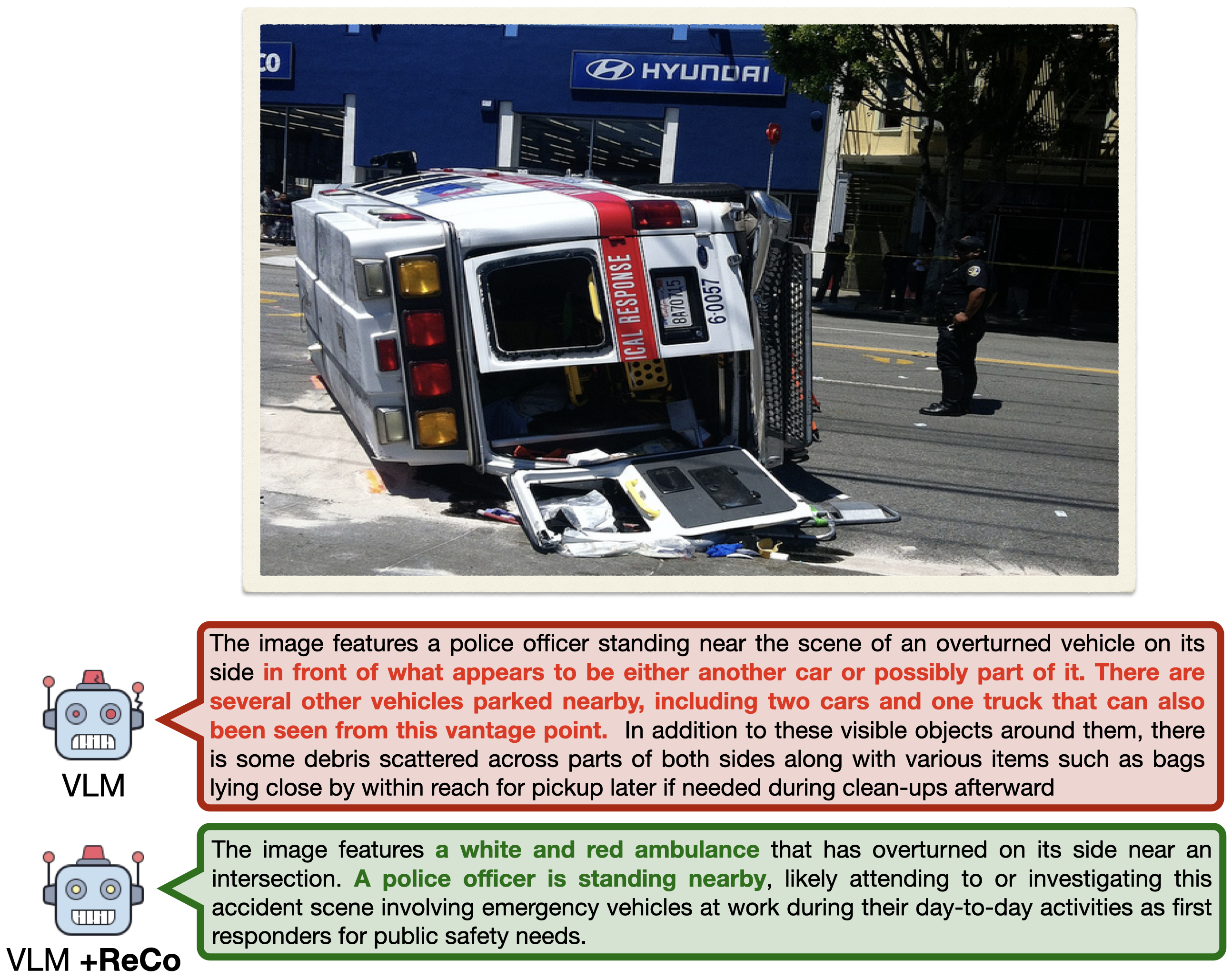}
    \caption{InstructBLIP \cite{instructblip} before and after ReCo. We propose a small module that, with minimal training, it is able to effectively reduce the hallucination rate of widely used VLMs.}
    \label{fig:instructblip_teaser}
    \vspace{-25pt}
\end{wrapfigure}

Given the  advances in the capabilities of Large Language Models (LLMs),  recent efforts have sought 
to extend these models to the multi-modality setting, i.e., processing and ``understanding''
additional modalities beyond text such as audio, images, and videos \cite{instructblip, llava1, qwen_audio, audiochatllama, videollama}. 
To this end, one milestone is the development of Vision Language Models (VLMs) 
that can accept both images and natural language as input, and generate contextually meaningful outputs for tasks, including visual question answering and image captioning.
Some prominent models are InstructBLIP \cite{instructblip}, MiniGPT4 \cite{minigpt4,minigptv2} and Llava \cite{llava1,llava2}, that show strong \textit{image+text} understanding skills. 
However, we know that VLMs piggyback 
heavily on the core capabilities of the parent LLM to which the visual representations have been aligned. This endows sizable compute benefits -- 
InstructBLIP \cite{instructblip} costs about $500$ GPU hours, while Llama \cite{llama} (the parent LLM), needed $180000$ GPU hours. 
But this VLM/LLM dependence means that VLMs also inherit known weaknesses of LLMs and sometimes, these weaknesses can be magnified. One example is hallucination, i.e., generating text that is plausible but does not accurately reflect the provided input. 
More than a handful of results in the literature show that in many cases, a VLM 
appears to ignore the image and generates a  description that is not influenced much at all by this extra visual input \cite{m3id, vcd, avisc}, although this is an extreme case of hallucination. 
We show an example in \Cref{fig:instructblip_teaser} where InstructBLIP \cite{instructblip} ``sees'' multiple cars and trucks, although we only see an ambulance.

\begin{wrapfigure}{r}{0.43\textwidth}
    \centering
    \vspace{-12pt}
\includegraphics[width=\linewidth]{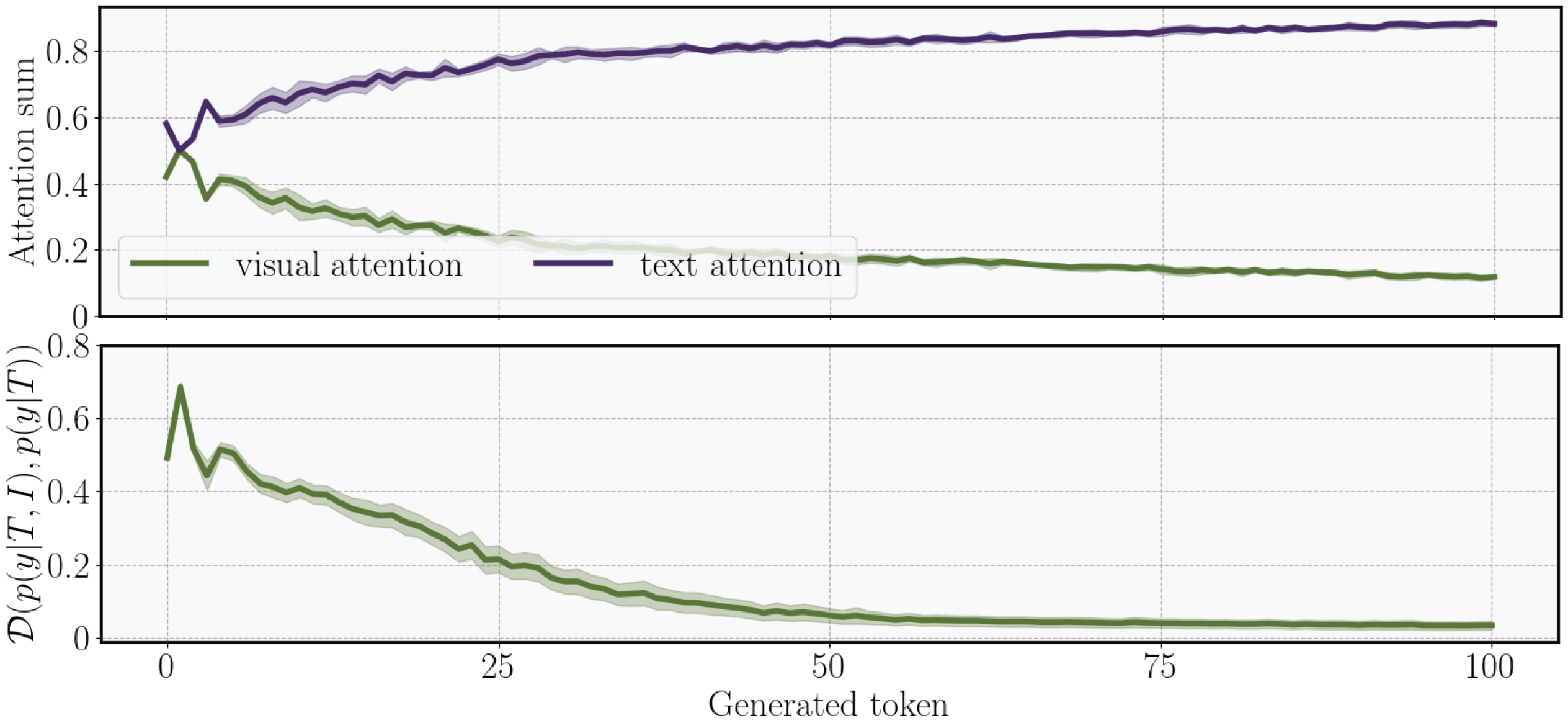}
    \caption{The ``fading memory effect''. \textbf{Top}: First-layer attention of the visual and textual input as the generation progresses. 
    %The model tends to ``forget'' about the visual input and focuses solely on the generated text. 
    \textbf{Bottom}: The effect of the visual input on the logits distribution as the generation progresses. We calculate the next token prediction with and without the visual input and we compute the distributional difference. After the first tokens, the next token can be predicted just from the previously generated text.}
    \label{fig:fading_memory}
    \vspace{-15pt}
\end{wrapfigure}

\noindent\textbf{Available mitigation mechanisms?} Mitigation mechanisms for the foregoing 
problem 
can be classified into two categories: 
\begin{inparaenum}[\bfseries (i)]
    \item train-based methods, where the VLM is further finetuned with different loss functions and/or using more suitable datasets \cite{dpo, hadpo, hallucidoctor, hacl}; and 
    \item rule-based methods, where the VLM remains frozen but a new/improved 
    generation process must be adopted. Some 
    proposals treat the model as a black box without any access to its attention maps and its parameters \cite{m3id, vcd}, whereas others use them to steer the model's generation process \cite{opera, avisc, vti}. 
\end{inparaenum}

\noindent\textbf{Why do VLMs hallucinate?} The literature suggests that VLMs over-rely on language priors, gradually ``forgetting" the visual input as text generation progresses \cite{m3id, avisc, opera, hacl, vcd}. 
If next token 
generation is implicitly 
probing an internal conditional probability distribution, 
$\mathcal{P}(\cdot|\circlerighthalfblack$), where the conditioning is with respect to both 
the text $T_t$ and the image $I$ with the right balance $\circlerighthalfblack$, progressively 
the image's role diminishes 
to $\mathcal{P}(\cdot|\circleurquadblack)$
and finally to $\mathcal{P}(\cdot|\mdwhtcircle)$, 
an effect that \cite{m3id} calls the \textit{``fading memory effect''}. 
We further probe this behavior: in \cref{fig:fading_memory} (top), we can observe that during the generation process, the total attention on the visual tokens drops significantly, showing that the model tends to ``ignore'' this information and rely on the language prior as it generates more and more text. In \cref{fig:fading_memory} (bottom), following \cite{m3id}, we compute the difference in the logits distribution with/without the image. If $T_t$ is the text generated so far and $I$ is the visual input, we calculate the difference between the probability distributions $\mathcal{P}(y_{t+1}|T_t, I)$ and $\mathcal{P}(y_{t+1}|T_t)$ where $y_{t+1}$ denotes the next token prediction. 
Using the Hellinger distance,  we see that the distance drops to almost $0$ after the first $40$ tokens: the image has practically no effect on the generated tokens afterwards.

\begin{wrapfigure}{r}{0.45\textwidth}
    \centering
    \vspace{-16pt}
    \includegraphics[width=\linewidth]{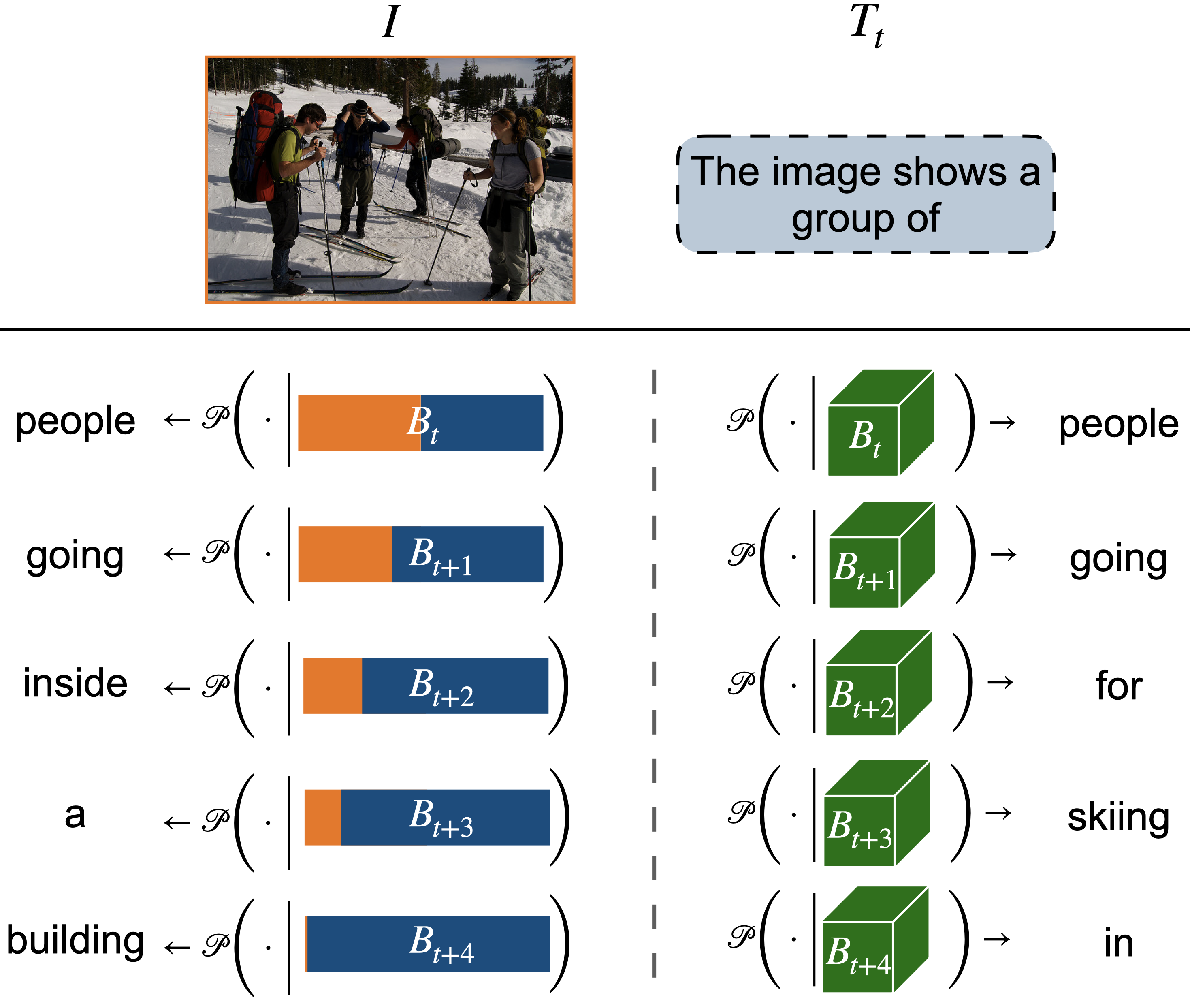}
    \caption{VLM: ideal versus practice. On the left, we show what actually happens, where at each timestep the image's influence  ($I$) in orange is diminished compared to the text $(T_t)$ in blue. So, the generated text is not an accurate representation of the visual input. An ideal VLM (right side) would form an object ($B_t$) that 
    perfectly encapsulates {\em all} of the given input, leading to accurate generation.}
    \label{fig:ideal}
    \vspace{-20pt}
\end{wrapfigure}

\noindent\textbf{The desired behavior.} A potential mitigation strategy will involve a ``correct'' composition of both the visual and the textual embeddings. But we should avoid overhauling the 
entire VLM architecture if possible.
%and instead forming 
%an explicit composition of the visual and textual embeddings prior to the next token generation.
%This observation is orthogonal to how a VLM should operate. 
If a VLM estimates the probability distribution $\mathcal{P}(y_{t+1}|B_t)$ where $B_t$ (shown as $\circlerighthalfblack$ above) encapsulates all the required information for the next token, we would expect that the quantity $B_t$ combines (or {\bf composes}) both $T_t$ and $I$, and is sensitive to changes in {\em any} of the two inputs for all $t$. However, this does not always happen in practice -- so we must intervene and carefully design $B_t$ so that neither of the inputs ``get lost'' (\cref{fig:ideal}).

\noindent\textbf{Compositionality and Geometric Algebra.} 
Both \begin{inparaenum} [\bfseries (i)]
    \item compositional learning and
    \item geometric algebra
\end{inparaenum}  are
mature topics of research \cite{attributes_operators, geometric1, geometric2}, and inform our approach.
{\em Compositionality} means that complex expressions are based on the meaning of their constituents and rules for combination. This was central to various visual understanding tasks: recognizing combinations of attributes and objects (e.g., ``spotted giraffe'' \cite{graph_compositional, open_world_compositions}), understanding object interactions (e.g., ``person holding umbrella'' \cite{visual_genome}), identifying transformations (e.g., ``broken glass'' \cite{abhinav_compositions}), and parsing complex scenes \cite{neuro_vqa}. 
Composition is not widely used as a standalone solution, but it is 
relevant in interpretability and checking 
semantic validity \cite{multi_compositions, hrr_compositions}. 
{\em Geometric algebra} generalizes concepts like complex numbers, quaternions, and vector algebra into a single framework for writing down geometric relationships/transformations. While early applications were in physics, the ability to manipulate geometric objects through algebraic operations like the geometric product (captures inner {\em and} outer product) is finding use in graphics \cite{geometric_graphics} and, more recently, in machine learning \cite{geometric_transformer, clifford_networks}.% chytas2024geometric}.

\noindent\textbf{Contributions.} This paper provides mitigation strategies for the fading memory effect by leveraging the compositional concepts above. We propose {\bf ReCo}, a lightweight module which is data-driven but treats the VLM as a black box, combining the best of both worlds. ReCo can be easily deployed on top of any VLM during/after training and, with minimal effort, improves their hallucination behavior. We achieve promising improvements on three widely used VLMs (InstructBLIP \cite{instructblip}, LlaVA \cite{llava1}, and MiniGPT4 \cite{minigpt4}) across multiple benchmarks and datasets (POPE \cite{pope}, CHAIR \cite{chair}, AMBER \cite{amber}, HallusionBench \cite{hallusionbench}, and MME \cite{mme}). 
%Our results show that, de
Despite the small size/compute footprint, 
%such an approach can provide 
we get a performance boost without any increase in the inference time of the VLMs. 
Furthermore, ReCo can be combined with virtually any of the rule-based methods without any modifications.
%, strengthening further the results across all benchmarks.

%\paragraph{Paper structure.} \cref{sec:background} provides the necessary background regarding VLMs and hallucinations as well as Geometric Algebra and Relational Compositions, the theoretical tools that guide our practical design choices. \cref{sec:method} details our approach (named ReCo) and the results are presented in \cref{sec:experiments}. Finally, \cref{sec:conclusion} provides the final remarks of our work.

%\section{Vision-Language Models and Hallucinations}\label{sec:hallucinations}
%\input{sec/hallucinations}

\section{Preliminaries: Geometric Algebra}\label{sec:background}

Geometric (or Clifford) Algebra (GA), an extension of linear algebra, deals with a ring $\mathcal{G}$, i.e., a set of elements $\mathcal{G}$ accompanied by two operations: $\oplus$ and $\otimes$ \cite{geometric1,geometric2}. The elements of GA (i.e., the elements of $\mathcal{G}$) are also vectors -- $\mathcal{G}$ contains scalars (or $0$-vectors), ``typical'' vectors (or $1$-vectors) as well as $2$-vectors, $3$-vectors, and so on. The subset of $\mathcal{G}$ that contains all $k$-vectors is usually denoted as $\mathcal{G}_k$. 
\newline\newline
\noindent\textbf{Multi-vectors.} A $k$-vector can be defined as the geometric product of $k$ orthogonal $1$-vectors, i.e., if $\{\mathbf{v}_1,\cdots,\mathbf{v}_k\}\in\mathcal{G}_1$ and $\mathbf{v}_i\perp\mathbf{v}_j,\ \forall i\neq j\in [k]\times [k]$, then 
\begin{equation}
%\small
    \mathbf{v}_1\otimes \cdots \otimes \mathbf{v}_k = \mathbf{v}^{(k)}\in\mathcal{G}_k
\end{equation}
Alternatively, each $k$-vector can be defined as the ``wedge'' ($\wedge$) product of $k$ $1$-vectors, where the wedge product is:
\begin{equation}
%\small
    \mathbf{v}_1\wedge \cdots \wedge \mathbf{v}_k = \frac{1}{k!}\bigoplus_{\sigma}\text{sign}(\sigma) \otimes\mathbf{v}_{\sigma(1)}\otimes\cdots \otimes\mathbf{v}_{\sigma(k)}, 
\end{equation}
where $\sigma$ denotes a permutation from the symmetric group. The sign is $+1$ (or $-1$) for even (or odd) permutations and the following equivalence holds:
\begin{equation}
%\small
    \mathbf{v}_1\otimes \cdots \otimes \mathbf{v}_k = \mathbf{v}_1\wedge \cdots \wedge \mathbf{v}_k \Leftrightarrow \mathbf{v}_1\perp \cdots \perp \mathbf{v}_k
\end{equation}

\noindent\textbf{Geometric operations.} A 2-vector represents a plane, a 3-vector a 3D object, and so on—unlike the inner product, which reduces vectors to a scalar with no recovery of inputs. The geometric product resembles and generalizes the outer product. These GA operations and multi-vectors motivate our instantiation: a systematic method to \textit{fuse} or \textit{compose} vectors beyond the common weighted averaging used in most architectures -- this is the relevance to our ``reminder composition''.
\cite{relational,attention,pooling}.
\newline\newline
\noindent\textbf{Practical implementations and limitations.}
While GA gives us the necessary axioms/properties, 
it does not provide a specific, practical instantiation. In general, a correct (but efficient) implementation in low dimensions ($\leq 6$) requires intensive 
effort, see \cite{lookmanomatrices}.
In general, since we deal with vectors, the typical choice would be the tensor product, which is infeasible for high dimensional vectors or multi-products (exponential memory), see \cite{geometric_graphics}, where quaternions 
are used for a computer graphics use case.
%, they are not applicable to our case since we deal with vectors of much higher dimensionality. For that reason, w
We shift our focus to ideas that can scale while preserving, to the extent possible, the general framework of GA.
\newline\newline
% \noindent\textbf{Vector Symbolic Architectures.} The result in \cite{relational} recently presented  ways to \textit{combine} multiple vectors together beyond linear combinations. Earlier works (e.g., \cite{hrr, vtb}) proposed practical instantiations under the name \textit{Vector Symbolic Architectures} (VSA) -- because one works with high-dimensional vectors but also, the two operations ($\otimes,\ \oplus$) resemble the bind and bundle operations of old results in symbolic or connectionist AI 
% \cite{logic_ai, connectionism, ai_bind, symbols_hinton}. More details 
% of different approaches and their properties  are in \cite{vsa}. 
% %Although these works focused on the symbolic aspects of AI, 
% We should note that some recent works \cite{fockbox} have linked these ideas to 
% concepts in physics. In our case, 
% %also demonstrated that there are deep connections between these architectures and mathematical spaces like the Fock Space and, as a result, with the Geometric Algebra itself, allowing us to 
% we simply ``borrow'' these operations for our application.
\noindent\textbf{Vector Symbolic Architectures.} \cite{relational} recently introduced ways to combine vectors beyond linear combinations. Earlier works \cite{hrr, vtb} formalized this under \textit{Vector Symbolic Architectures} (VSA), named for their use of high-dimensional vectors and operations ($\otimes,\ \oplus$) resembling bind and bundle in symbolic and connectionist AI \cite{logic_ai, connectionism, ai_bind, symbols_hinton}. For a detailed overview, see \cite{vsa}. Recent works \cite{fockbox} have linked these ideas to physics, particularly Fock Space, connecting them to Geometric Algebra. This allows us to adopt these operations directly for our application.
\newline\newline
\noindent\textbf{Relevance of these concepts.}
Our main hypothesis is that the \textit{fading memory effect} can be mitigated through a \textit{composition} of the textual and visual information. The logical question then is how this composition can be designed, say, first conceptually and then for practical use. Theoretically, GA provides us a structured framework for composition problems, enabling the integration of modalities such as image and text, {\bf without} the need of ad-hoc solutions. Practically, VSA offers efficient implementations with favorable computational complexity, memory usage, and noise tolerance while preserving GA’s properties.

\section{Reminder Composition (ReCo)}\label{sec:method}

%The goal of this work is to leverage the idea of multi-vectors and its practical instantiation, the relational compositions, 

{\bf Overview.} We outline our construction to minimize the \textit{fading memory effect}, by calculating $B_t$ as an explicit composition of $T_t$ and $I$ (\cref{fig:ideal}).
%Since the VLM models are of an autoregressive nature, the further they are down the generation process, the harder it is to ``keep'' all the required information for both the given image as well as for the so-far generated text into its next-token embedding. To frame it differently, \textit{{the model is not able to shape its latent space in a way that 
If the embedding $B_t$ can be designed 
to be an (almost) lossless multi-vector composition of the generated text and the input image, we will achieve improvements. 
%On the contrary, the image's influence is minimized, making its ``participation'' obsolete. To address this issue, 
Our simple idea is to explicitly form a proper multi-vector of the text and the image information {\em before} decoding the embedding to the token space.

\begin{wrapfigure}{r}{0.5\textwidth}
    \centering
    \vspace{-22pt}
    \includegraphics[width=\linewidth]{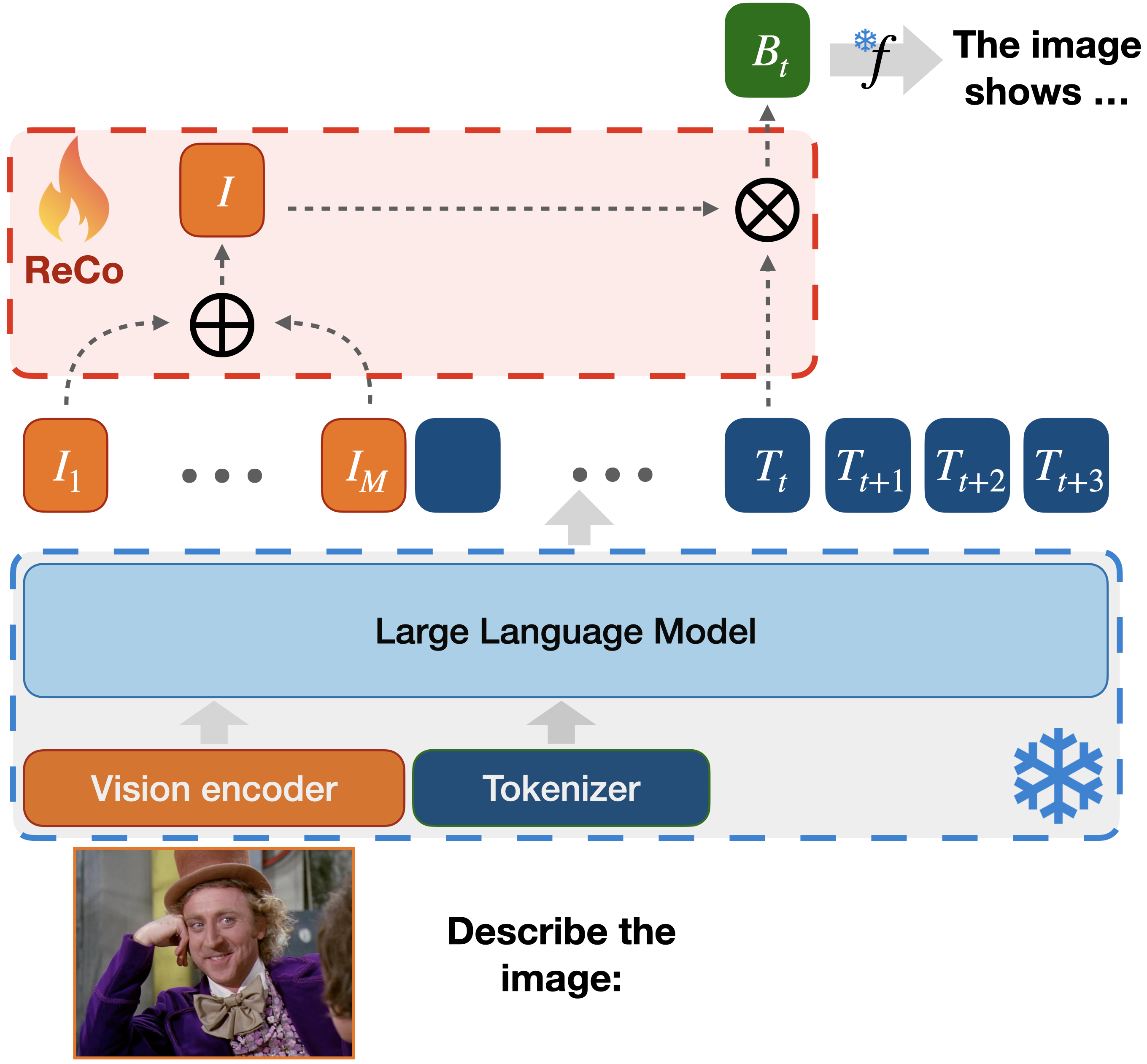}
    \caption{{\bf ReCo overview}. The VLM is treated as a black box, modifying the next token embedding by combining the multi-vector of visual tokens and the current token prediction. First, we bundle the image tokens $[I_j]_{j=1}^m$ into a single vector $I$ and, then, bind it with $T_t$ to form $B_t$, ensuring the image's influence. Finally, the prediction head $f$ (also frozen) predicts the next token $w_t$.}
    \label{fig:reco}
\end{wrapfigure}

Let $[T_t]_{t=1}^N$ be the VLM's predicted embeddings for each one of the $N$ generated tokens which are then fed into the LLM head ($f$) to form the next token $[w_t]_{t=1}^N$. $T_t$ corresponds to the hidden state of the LLM at the step $t$ (usually denoted as $\mathbf{h}_t$ but we use $T$ here to mnemonically suggest ``text''). Additionally, let $[I_j]_{j=1}^{M}$ be the $M$ output embeddings that correspond to the $M$ image embeddings that are fed to the VLM. An ideal VLM should closely capture the composition:
\begin{equation}
\small
    T_t = g\Bigg(T_{t-1}\otimes\bigg(\bigoplus_{j=1}^M I_j\bigg)\otimes \mathbf{p}_t\Bigg)
    \label{eq:comp}
\end{equation}
which states that the next token depends on the previous one {\em composed} with all of the provided image information, with $g$ denoting a learnable transformation of the multi-vector and $\mathbf{p}_i$ denoting some extra optional information about the index of the current token. The next token corresponds to $w_t = f(T_t)$ but now the image effect, by design, {\bf cannot} be neglected. Notice that (\ref{eq:comp}) simply re-formulates what we have already observed about $p(y_{t+1}|T_t, I)$ and $p(y_{t+1}|T_t)$ in \cref{fig:ideal}, showing how $B_t$ should actually behave in theory, i.e., 
\begin{align}
\begin{split}
\small
    B_t &= g\Bigg(T_{t-1}\otimes\bigg(\bigoplus_{j=1}^M I_j\bigg)\otimes \mathbf{p}_t\Bigg) \\
    w_t &= f(B_t)\ \ \propto \mathcal{P}\big(y_{t+1}|B_t\big)
\end{split}
\end{align}

\begin{wrapfigure}{r}{0.4\textwidth}
    \centering
    \includegraphics[width=\linewidth]{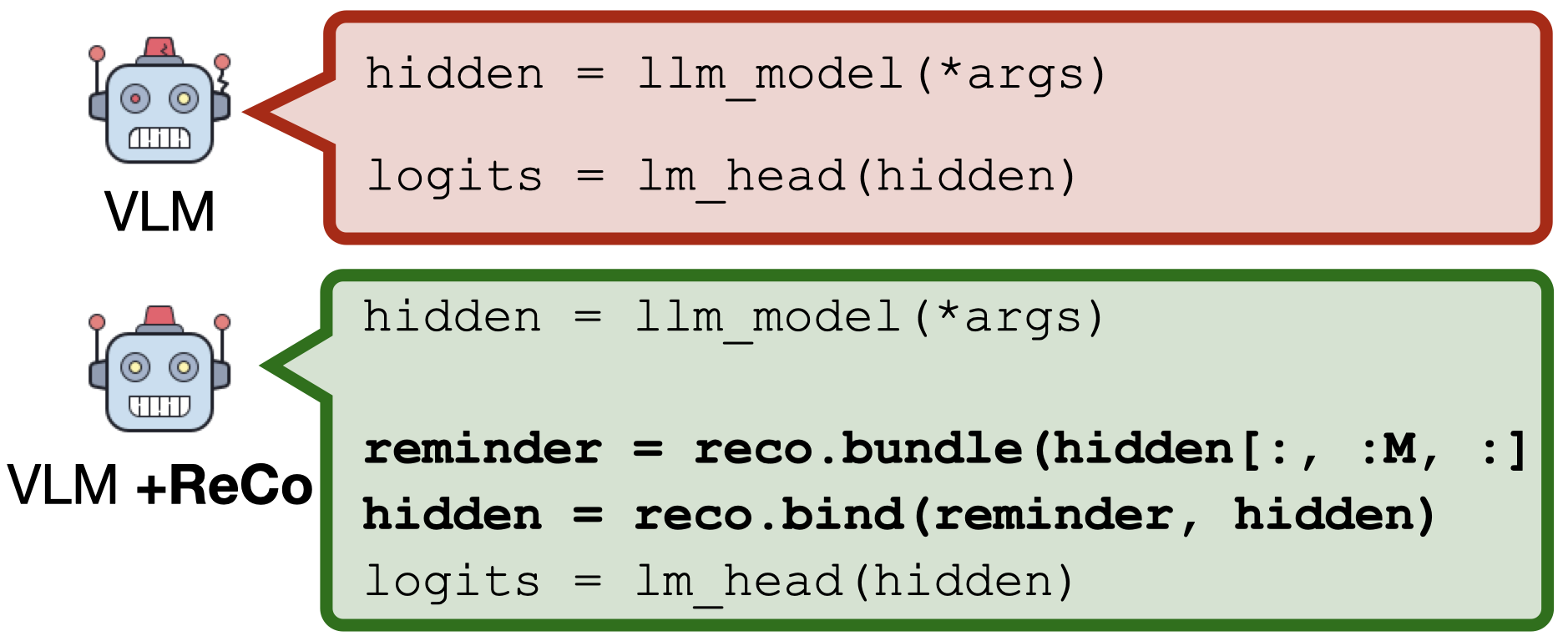}
    % \vspace{-20pt}
    \caption{ReCo in practice. No access to the LLM is required and the only change is the modification of the prediction head with the addition of a ``preprocessing'' step.}
    \label{fig:algo}
\end{wrapfigure}

Therefore, we propose to explicitly modify the VLM's output so that it corresponds to the composition as described in (\ref{eq:comp}). Based on \cite{relational}, we define the geometric product as the Matrix Binder operation, allowing us to mitigate 
the fading memory problem by adding only a small trainable layer on top of a frozen, black-box VLM as:
\begin{equation}
    \small
    B_t = W_T T_{t} +W_I\Bigg(\bigoplus_{j=1}^M I_j\Bigg)
    \label{eq:final}
\end{equation}
The modifications in any VLM's codebase consist of the addition of {\bf only two extra lines of code} without any other changes to the model's ``internals'' (see \cref{fig:algo}).

% \begin{lstlisting}
% hidden = llm_model(*args)

% # typical VLM
% logits = lm_head(hidden)

% # ReCo
% reminder = reco.bundle(hidden[:, :M, :])
% hidden = reco.bind(reminder, hidden)
% logits = lm_head(hidden)
% \end{lstlisting}

Many  composition rules have been proposed in the literature but our choice above was driven by two reasons:
\begin{asparaenum}[\bfseries (a)]
    \item The matrices $W_T, W_i$ can be trained alongside the VLM, allowing our modification to be integrated into any model without altering the training process while, at the same time, leveraging the explicit composition rule during inference and generation.%The matrices $W_T, W_i$ can be trained on the data that a VLM is trained on. This means that our modification can be deployed to any VLM during training without any further changes in the training procedure while, at the same time, taking advantage of the explicit composition rule during inference/generation.
    \item Our modified VLM extends the original model. Setting $W_T=I$ and $W_I=0$ restores the original VLM, making the modified solution space a {\em strict superset}. Thus, using ReCo in training retains all original capabilities while potentially improving hallucination mitigation.%Our modified VLM is a direct extension of the unmodified version. Notice that by setting $W_T=I$ and $W_I=0$, we recover the original VLM, which means that the solution of the modified VLM is a strict superset of the original solution space. This implies that, if one were to use ReCo during training, the modified VLM would not lose any of the capabilities of the unmodified one, while, potentially, achieving better results for mitigating hallucinations.
\end{asparaenum}
We should note that our formulation depends on a specific composition strategy. We used one which is mathematically sound but also efficient, but the operations can be upgraded.

\noindent\textbf{VLM is a black-box.} Notice that (\ref{eq:final}) and the corresponding code (\cref{fig:algo}) involves only the output layer of the VLM. This means that, in practice, no access to the model is needed. The output embeddings of the model can be calculated offline, and then we train only the few extra parameters of ReCo in a matter of minutes on any commodity GPU, without even needing to load the entire VLM in memory and performing multiple inference passes through it. This allows us to benefit from additional training data and more suitable training tactics (similar to fine-tuning approaches, e.g., \cite{hadpo}) while treating the VLM as a frozen black box (similar to decoding strategies, e.g., \cite{avisc, m3id}). A full comparison with existing solutions is in the appendix.
% \newline\newline
% \noindent\textbf{Avoiding overfitting and catastrophic forgetting.} Unless ReCo is part of the VLM training, it can be trained using 
% only a fraction of the dataset that
% was used for the training the full model. 
% In our experiments, to avoid overfitting to the small training dataset, we find it useful to use a positive weight for the ReCo term, essentially predicting the next token based on a weighted combination of the new multi-vector and the old hidden state, i.e.,
% \begin{equation}
% \small
%     B_t = \lambda \Bigg( T_t \big\otimes \bigg(\bigoplus_j I_j\bigg) \Bigg) + T_t
% \end{equation}
% where $\lambda\in(0, 1)$.

\section{Experiments}\label{sec:experiments}

Before diving into the technical details and comprehensive analysis,, we highlight some of our main findings which we will analyze in detail shortly. These results offer a high-level overview of the most significant outcomes of our work, which will be examined and discussed in depth below.

\begin{asparaenum}[\bfseries (a)]
    \item ReCo leads to noticeable improvements across \textbf{all} of the VLMs we evaluated. In every case, it effectively reduces the rate of hallucinations, demonstrating its robustness and general applicability.
    \item ReCo works seamlessly with other methods for reducing hallucinations, making it easy to integrate into different systems. Further, when combined with other approaches, it leads to even better performance, showing that the benefits of ReCo and other techniques can complement each other. 
    \item ReCo enables the model to effectively “recover” from an initial hallucination, distinguishing it from baseline VLMs which often persist in elaborating on hallucinated content. This ability to self-correct contributes to more coherent and accurate model outputs. 
    \item Our quantitative and qualitative analyses show that ReCo goes beyond just fixing object-related hallucinations. It also helps correct mistakes related to the overall structure of the image and the specific attributes or features of the objects themselves. 
\end{asparaenum}

\begin{figure*}
    \centering
    \includegraphics[width=\linewidth]{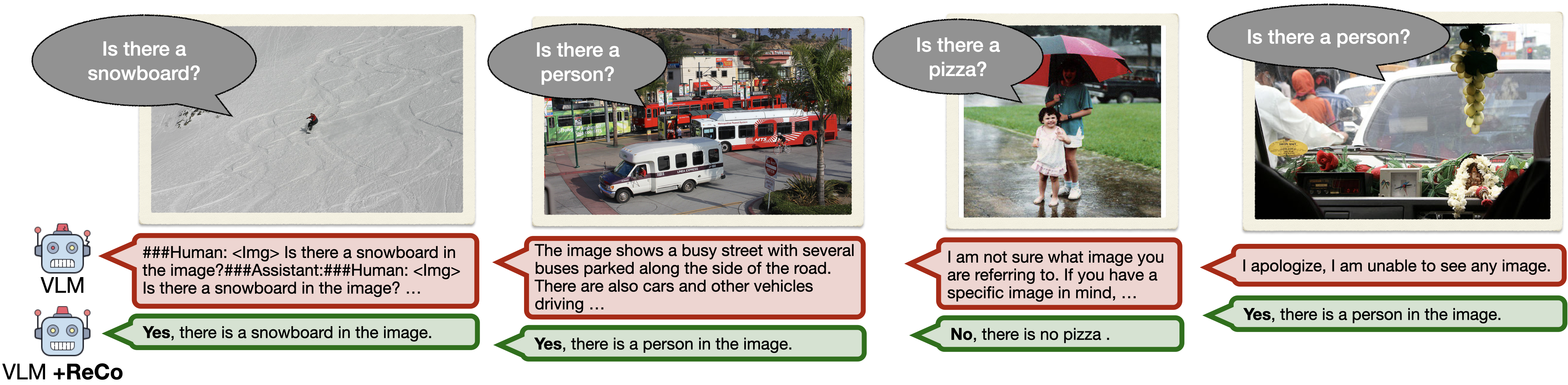}
    % \vspace{-10pt}
    \caption{POPE \cite{pope} results on MiniGPT4 \cite{minigpt4}. The unmodified version is often unable to comprehend the question and outputs an unrelated answer that does no contain either ``Yes'' or ``No''. On the contrary, ReCo provides the model the ability to answer such questions.}
    \label{fig:pope}
\end{figure*}

\setlength{\tabcolsep}{5pt} % Default value: 6pt

\begin{table*}[h]
    \scriptsize
    \centering
    \caption{CHAIR \cite{chair} results for both InstructBLIP \cite{instructblip} and MiniGPT4 \cite{minigpt4}. \textit{Original} stands for the unmodified VLM, while VCD \cite{vcd}, M3ID \cite{m3id}, and AvisC \cite{avisc} are the three baselines we consider. In all four models, the addition of ReCo improves significantly the performance as the generation progresses, reducing CHAIR$_\text{s}$ as much as $44\%$ and CHAIR$_\text{i}$ $30\%$, with $32,\ 64,\ 128,\ 256,\ 512,\ \text{and } 1024$ standing for the maximum allowed number of generated tokens.}
    \label{tab:chair}
    \begin{NiceTabular}{cl|c c c c c c | c c c c c c | c }
         \toprule
          \multicolumn{2}{l|}{\multirow{2}{*}{Model}} & \multicolumn{6}{c}{CHAIR$_\text{s}(\downarrow)$} & \multicolumn{6}{c}{CHAIR$_\text{i}(\downarrow)$} & \multirow{2}{*}{Average length} \\
          \cmidrule(lr){3-8}\cmidrule(lr){9-14}
          &  & $32$ & $64$ & $128$ & $256$ & $512$ & $1024$ & $32$ & $64$ & $128$ & $256$ & $512$ & $1024$ & \\
          \midrule
         \multirow{8}*{\rotatebox[origin=c]{90}{InstructBLIP \cite{instructblip}}}& Original & $8.6$ & $22.0$ & $37.4$ & $37.2$ & $37.2$ & $37.2$ & $4.8$ & $8.3$ & $11.8$ & $12.2$ & $12.2$ & $12.2$ & $456$ \\
         & \textbf{+ReCo} & \textcolor{NEGATIVE}{+$0.8$} & \textcolor{NEGATIVE}{+$0.2$} & \textcolor{POSITIVE}{-$7.0$} & \textcolor{POSITIVE}{-$8.8$} & \textcolor{POSITIVE}{-$8.8$} &
         \textcolor{POSITIVE}{-$8.8$} & \textcolor{NEGATIVE}{+$0.5$} & \textcolor{NEGATIVE}{+$0.3$} & \textcolor{POSITIVE}{-$1.1$} & \textcolor{POSITIVE}{-$1.6$} &
         \textcolor{POSITIVE}{-$1.6$} & \textcolor{POSITIVE}{-$1.6$}  & $350$ \\
         % \tikzdashline{2}{15}\\[-2ex]
         % \booktabdashline
         \cdottedline{2-15}
         % \cdashline{2-15}[2pt/3pt]\\[-2ex]
         & VCD \cite{vcd} & $9.6$ & $19.2$ & $37.6$ & $36.0$ & $36.0$ & $36.0$ & $5.0$ & $6.7$ & $11.2$ & $11.4$ & $11.4$ & $11.4$ & $427$  \\
         & \textbf{+ReCo} & \textcolor{POSITIVE}{-$1.0$} & \textcolor{POSITIVE}{-$0.8$} & \textcolor{POSITIVE}{-$8.6$} & \textcolor{POSITIVE}{-$8.6$} & \textcolor{POSITIVE}{-$8.6$} & \textcolor{POSITIVE}{-$8.6$} & \textcolor{POSITIVE}{-$0.5$} & \textcolor{NEGATIVE}{+$0.5$} & \textcolor{POSITIVE}{-$1.5$} & \textcolor{POSITIVE}{-$2.8$} & \textcolor{POSITIVE}{-$2.8$} & \textcolor{POSITIVE}{-$2.8$}  & $329$ \\
         \cdottedline{2-15}
         % \cdashline{2-15}[2pt/3pt]\\[-2ex]
         & M3ID \cite{m3id} & $8.6$ & $22.2$ & $31.2$ & $32.8$ & $32.8$ & $32.8$ & $3.9$ & $7.1$ & $9.7$ & $9.6$ & $9.6$ & $9.6$  & $410$  \\
         & \textbf{+ReCo} & \textcolor{POSITIVE}{-$2.0$} & \textcolor{POSITIVE}{-$2.6$} & \textcolor{POSITIVE}{-$5.8$} & \textcolor{POSITIVE}{-$7.4$} & \textcolor{POSITIVE}{-$7.4$} & \textcolor{POSITIVE}{-$7.4$} & \textcolor{POSITIVE}{-$0.8$} & \textcolor{POSITIVE}{-$0.7$} & \textcolor{POSITIVE}{-$1.3$} & \textcolor{POSITIVE}{-$1.2$} & \textcolor{POSITIVE}{-$1.2$} & \textcolor{POSITIVE}{-$1.2$} & $294$\\
         \cdottedline{2-15}
         % \cdashline{2-15}[2pt/3pt]\\[-2ex]
         & AvisC \cite{avisc} & $8.2$ & $17.6$ & $32.2$ & $33.0$ & $33.0$ & $33.0$ & $4.2$ & $6.4$ & $9.3$ & $9.1$ & $9.1$  & $9.1$ & $408$ \\
         & \textbf{+ReCo} & \textcolor{POSITIVE}{-$2.2$} & \textcolor{NEGATIVE}{+$0.2$} & \textcolor{POSITIVE}{-$9.2$} & \textcolor{POSITIVE}{-$8.8$} & \textcolor{POSITIVE}{-$8.8$} & \textcolor{POSITIVE}{-$8.8$} & \textcolor{POSITIVE}{-$1.5$} & \textcolor{POSITIVE}{-$0.4$} & \textcolor{POSITIVE}{-$2.2$} & \textcolor{POSITIVE}{-$1.4$} & \textcolor{POSITIVE}{-$1.4$} & \textcolor{POSITIVE}{-$1.4$} & $311$ \\
         \toprule
         \multirow{8}*{\rotatebox[origin=c]{90}{MiniGPT4 \cite{minigpt4}}}& Original & $9.6$ & $19.4$ & $31.4$ & $36.2$ & $37.6$ & $37.6$ & $4.9$ & $7.5$ & $9.9$ & $11.7$ & $12.8$ & $13.4$ & $374$ \\
         & \textbf{+ReCo} & \textcolor{NEGATIVE}{+$2.6$} & \textcolor{NEGATIVE}{+$2.2$} & \textcolor{POSITIVE}{-$2.8$} & \textcolor{POSITIVE}{-$4.6$} & \textcolor{POSITIVE}{-$5.4$} & \textcolor{POSITIVE}{-$4.8$} & \textcolor{NEGATIVE}{+$1.1$} & \textcolor{NEGATIVE}{+$1.3$} & \textcolor{POSITIVE}{-$0.4$} & \textcolor{POSITIVE}{-$1.5$} & \textcolor{POSITIVE}{-$2.6$} & \textcolor{POSITIVE}{-$3.2$} & $307$ \\
         \cdottedline{2-15}
         % \cdashline{2-15}[2pt/3pt]\\[-2ex]
         & VCD \cite{vcd} & $10.0$ & $20.8$ & $31.0$ & $35.8$ & $36.6$ & $36.6$ & $5.1$ & $8.5$ & $10.5$ & $12.1$ & $13.0$ & $13.5$ & $336$  \\
         & \textbf{+ReCo} & \textcolor{NEGATIVE}{+$0.8$} & \textcolor{POSITIVE}{-$2.2$} & \textcolor{POSITIVE}{-$7.6$} & \textcolor{POSITIVE}{-$9.6$} & \textcolor{POSITIVE}{-$8.8$} & \textcolor{POSITIVE}{-$8.4$} & \textcolor{NEGATIVE}{+$0.1$} & \textcolor{POSITIVE}{-$0.4$} & \textcolor{POSITIVE}{-$0.7$} & \textcolor{POSITIVE}{-$1.2$} & \textcolor{POSITIVE}{-$1.5$} & \textcolor{POSITIVE}{-$1.9$} & $249$ \\
         \cdottedline{2-15}
         % \cdashline{2-15}[2pt/3pt]\\[-2ex]
         & M3ID \cite{m3id} & $9.4$ & $22.2$ & $39.2$ & $48.5$ & $49.5$ & $49.5$ & $4.6$ & $9.1$ & $12.3$ & $15.5$ & $15.9$ & $16.0$ & $372$\\
         & \textbf{+ReCo} & \textcolor{NEGATIVE}{+$2.6$} & \textcolor{NEGATIVE}{+$3.2$} & \textcolor{POSITIVE}{-$2.2$} & \textcolor{POSITIVE}{-$9.1$} & \textcolor{POSITIVE}{-$9.9$} & \textcolor{POSITIVE}{-$9.9$} & \textcolor{NEGATIVE}{+$1.0$} & \textcolor{NEGATIVE}{+$0.6$} & \textcolor{NEGATIVE}{+$0.2$} & \textcolor{POSITIVE}{-$1.4$} & \textcolor{POSITIVE}{-$1.6$} & \textcolor{POSITIVE}{-$1.7$} & $380$\\
         \cdottedline{2-15}
         % \cdashline{2-15}[2pt/3pt]\\[-2ex]
         & AvisC \cite{avisc} & $10.8$ & $20.8$ & $30.8$ & $41.4$ & $46.4$ & $48.2$ & $5.2$ & $7.9$ & $10.1$ & $12.1$ & $14.6$ & $16.0$ & $277$ \\
         & \textbf{+ReCo} & \textcolor{POSITIVE}{-$1.8$} & \textcolor{POSITIVE}{+$4.4$} & \textcolor{POSITIVE}{-$8.6$} & \textcolor{POSITIVE}{-$17.2$} & \textcolor{POSITIVE}{-$20.8$} & \textcolor{POSITIVE}{-$22.2$} & \textcolor{POSITIVE}{-$0.4$} & \textcolor{POSITIVE}{-$1.1$} & \textcolor{POSITIVE}{-$2.1$} & \textcolor{POSITIVE}{-$3.1$} & \textcolor{POSITIVE}{-$4.5$} & \textcolor{POSITIVE}{-$5.1$} & $220$ \\
         \toprule
    \end{NiceTabular}
    \vspace{-16pt}
    
\end{table*}

\setlength{\tabcolsep}{6pt} % Default value: 6pt

\subsection{Experimental setup}

\paragraph{Datasets and Training.} We use the same training procedure across all experiments, training ReCo with HA-DPO \cite{hadpo} while keeping the rest of the model frozen, unlike native HA-DPO, which finetunes the entire VLM. The dataset consists of quadruples ($I$, $P$, $C$, $R$), where $I$ is the image, $P$ is its associated prompt or question, and $C$ and $R$ are the \textit{Chosen} and \textit{Rejected} answers for the Direct Preference Optimization \cite{dpo}. The images are a small subset of Visual Genome \cite{visual_genome} (only $1853$ in total), with $C$ and $R$ generated using GPT-4 through a three-stage process \cite{hadpo}.
\newline\newline
\noindent\textbf{Evaluation.} To systematically evaluate ReCo, we use five benchmarks: CHAIR \cite{chair}, POPE \cite{pope}, AMBER \cite{amber}, HallusionBench \cite{hallusionbench}, and MME \cite{mme}.\begin{inparaenum}[\bfseries (a)] \item POPE asks binary existence questions like \textit{“Is there a $\langle$object$\rangle$ in the image?”}.
\item CHAIR measures hallucination rates in image captioning.
\item AMBER assesses both discriminative and generative capabilities with binary and open-ended questions such as \textit{“Is the umbrella open?”} and \textit{“Describe the image.”}.
\item HallusionBench and MME test various discriminative questions, evaluating accuracy and accuracy$+$ (see appendix).
\end{inparaenum}
CHAIR \cite{chair} and AMBER \cite{amber} use MSCoco \cite{mscoco}, while POPE \cite{pope} is based on MSCoco, A-OKVQA \cite{aokvqa}, and GQA \cite{gqa}. 

\paragraph{Baselines.} Beyond comparing with unmodified VLMs (InstructBLIP \cite{instructblip}, MiniGPT4 \cite{minigpt4}, and LlaVA \cite{llava1}), we evaluate three recent hallucination mitigation methods: M3ID \cite{m3id}, VCD \cite{vcd}, and AvisC \cite{avisc}. We report their performance before and after integration with ReCo to assess complementarity. Due to space constraints and the similar performance of InstructBLIP and LlaVA, detailed LlaVA results are provided in the appendix.

\subsection{Results}

Our extensive experiments show that ReCo can be deployed easily with any VLM and work well across multiple benchmarks. Additionally, it can be combined efficiently with other methods to strengthen the results further. Below, we analyze the results on each benchmark separately.

\subsubsection{CHAIR: Experimental evaluations} We prompt the model with ``Describe the image.'', without providing any additional information about the length of the description (e.g., ``Provide a short description of the image''). In \cref{tab:chair} (and Tab. 4 in the appendix), we present the improvement we achieve for both CHAIR$_\text{s}$ and CHAIR$_\text{i}$ metrics as we increase the length of the generated response.

ReCo helps -- reducing CHAIR$\text{s}$ by up to $44\%$ and CHAIR$\text{i}$ by $30\%$. While baselines sometimes perform better with fewer tokens, this is often due to extra characters before the actual text, affecting the effective output length. The key trend is that as token count increases, ReCo consistently improves both metrics {\em significantly}.

{\bf Analysis.} Like other mitigation methods, ReCo cannot fully eliminate hallucinations. However, its impact is evident in the significant improvement on CHAIR$_\text{s}$. While hallucinations may still occur, ReCo prevents the model from fixating and building upon them. For example, in \cref{fig:instructblip_teaser}, a model might mistakenly generate \textit{intersection}, but unlike the unmodified VLM, it does not continue elaborating on this error. This is due to ReCo’s image ``reminder'', which helps steer generation back toward accuracy, counteracting the over-reliance on language priors observed in VLMs (\cref{fig:fading_memory}).

\setlength{\tabcolsep}{3.8pt} % Default value: 6pt

\begin{table*}
    \scriptsize
    \centering
    \captionof{table}{POPE \cite{pope} results for MSCoco \cite{mscoco} and A-OKVQA \cite{aokvqa}. In all cases, ReCo provides a significant performance boost to all the methods (where \textit{Original} stands for the unmodified VLM).}
      \label{tab:pope}
    
    \begin{NiceTabular}{cl|cccccc|cccccc}
         \toprule
         & & \multicolumn{6}{c|}{MSCoco \cite{mscoco}} & \multicolumn{6}{c}{A-OKVQA \cite{aokvqa}} \\
         \cmidrule(lr){3-8}\cmidrule(lr){9-14}
         \multicolumn{2}{l|}{\multirow{1}{*}{Model}} & \multicolumn{2}{c}{Random} & \multicolumn{2}{c}{Popular}  & \multicolumn{2}{c|}{Adversarial} & \multicolumn{2}{c}{Random} & \multicolumn{2}{c}{Popular}  & \multicolumn{2}{c}{Adversarial} \\
          \cmidrule(lr){3-4}\cmidrule(lr){5-6}\cmidrule(lr){7-8}\cmidrule(lr){9-10}\cmidrule(lr){11-12}\cmidrule(lr){13-14}
          &  & Acc ($\uparrow$) & F1 ($\uparrow$) & Acc ($\uparrow$) & F1 ($\uparrow$)  & Acc ($\uparrow$) & F1 ($\uparrow$) & Acc ($\uparrow$) & F1 ($\uparrow$) &Acc ($\uparrow$) & F1 ($\uparrow$)  & Acc ($\uparrow$) & F1 ($\uparrow$) \\
         % \hline 
         % \hline\\[-2ex]
         \midrule
         \multirow{8}*{\rotatebox[origin=c]{90}{InstructBLIP \cite{instructblip}}}& Original & $82.8\%$ & $82.8\%$ & $77.6\%$ & $79.3\%$ & $74.6\%$ & $76.8\%$ & $80.6\%$ & $82.0\%$ & $73.9\%$ & $77.4\%$ & $67.8\%$ & $73.3\%$ \\
         & \textbf{+ReCo} & \textcolor{POSITIVE}{+$1.3\%$} & \textcolor{POSITIVE}{+$0.3\%$} & \textcolor{POSITIVE}{+$1.5\%$} & \textcolor{NEGATIVE}{-$0.6\%$} & \textcolor{POSITIVE}{+$2.8\%$} & \textcolor{POSITIVE}{+$0.5\%$} & \textcolor{POSITIVE}{+$3.4\%$} & \textcolor{POSITIVE}{+$0.5\%$} & \textcolor{POSITIVE}{+$6.9\%$} & \textcolor{POSITIVE}{+$2.2\%$} & \textcolor{POSITIVE}{+$8.1\%$} & \textcolor{POSITIVE}{+$2.1\%$}  \\
         \cdottedline{2-14}
         % \cdashline{2-14}[2pt/3pt]\\[-2ex]
         & VCD\cite{vcd} & $83.2\%$ & $83.2\%$ & $77.6\%$ & $78.9\%$ & $75.8\%$ & $77.5\%$ & $82.0\%$ & $83.2\%$ & $74.9\%$ & $77.9\%$ & $69.7\%$ & $74.6\%$  \\
         & \textbf{+ReCo} & \textcolor{POSITIVE}{+$1.6\%$} & \textcolor{POSITIVE}{+$0.2\%$} & \textcolor{POSITIVE}{+$2.9\%$} & \textcolor{POSITIVE}{+$0.8\%$} & \textcolor{POSITIVE}{+$3.6\%$} & \textcolor{POSITIVE}{+$1.3\%$} & \textcolor{POSITIVE}{+$3.2\%$} & \textcolor{POSITIVE}{+$0.3\%$} & \textcolor{POSITIVE}{+$8.0\%$} & \textcolor{POSITIVE}{+$3.7\%$} & \textcolor{POSITIVE}{+$7.1\%$} & \textcolor{POSITIVE}{+$1.7\%$}  \\
         \cdottedline{2-14}
         % \cdashline{2-14}[2pt/3pt]\\[-2ex]
         % \cmidrule(lr){2-8}
         & M3ID\cite{m3id} & $84.1\%$ & $84.2\%$ & $77.9\%$ & $79.3\%$ & $75.0\%$ & $77.2\%$ & $82.7\%$ & $83.8\%$ & $75.9\%$ & $78.7\%$ & $67.9\%$ & $73.6\%$  \\
         & \textbf{+ReCo} & \textcolor{POSITIVE}{+$1.1\%$} & \textcolor{NEGATIVE}{-$0.3\%$} & \textcolor{POSITIVE}{+$3.0\%$} & \textcolor{POSITIVE}{+$0.9\%$} & \textcolor{POSITIVE}{+$4.9\%$} & \textcolor{POSITIVE}{+$1.3\%$} & \textcolor{POSITIVE}{+$3.1\%$} & \textcolor{POSITIVE}{+$0.4\%$} & \textcolor{POSITIVE}{+$6.8\%$} & \textcolor{POSITIVE}{+$2.5\%$} & \textcolor{POSITIVE}{+$9.2\%$} & \textcolor{POSITIVE}{+$3.1\%$}  \\
         \cdottedline{2-14}
         % \cdashline{2-14}[2pt/3pt]\\[-2ex]
         & AvisC\cite{avisc} & $88.3\%$ & $87.8\%$ & $81.9\%$ & $82.3\%$ & $79.5\%$ & $80.3\%$ & $86.2\%$ & $86.7\%$ & $80.4\%$ & $82.1\%$ & $71.5\%$ & $75.8\%$  \\
         & \textbf{+ReCo} & \textcolor{NEGATIVE}{-$0.9\%$} & \textcolor{NEGATIVE}{-$1.2\%$} & \textcolor{POSITIVE}{+$1.7\%$} & \textcolor{POSITIVE}{+$0.7\%$} & \textcolor{POSITIVE}{+$2.2\%$} & \textcolor{POSITIVE}{+$1.2\%$} & \textcolor{POSITIVE}{+$1.1\%$} & \textcolor{NEGATIVE}{-$0.6\%$} & \textcolor{POSITIVE}{+$3.0\%$} & \textcolor{POSITIVE}{+$0.5\%$} & \textcolor{POSITIVE}{+$5.5\%$} & \textcolor{POSITIVE}{+$1.6\%$}  \\
        \midrule
        \multirow{8}*{\rotatebox[origin=c]{90}{MiniGPT4 \cite{minigpt4}}}& Original & $55.9\%$ & $36.1\%$ & $53.3\%$ & $34.8\%$ & $53.4\%$ & $34.8\%$ & $55.3\%$ & $36.6\%$ & $55.7\%$ & $33.0\%$ & $55.3\%$ & $32.8\%$  \\
         & \textbf{+ReCo} & \textcolor{POSITIVE}{+$2.0\%$} & \textcolor{POSITIVE}{+$13.9\%$} & \textcolor{POSITIVE}{+$3.1\%$} & \textcolor{POSITIVE}{+$14.3\%$} & \textcolor{POSITIVE}{+$2.2\%$} & \textcolor{POSITIVE}{+$9.8\%$} & \textcolor{POSITIVE}{+$6.8\%$} & \textcolor{POSITIVE}{+$7.4\%$} & \textcolor{POSITIVE}{+$6.1\%$} & \textcolor{POSITIVE}{+$10.8\%$} & \textcolor{POSITIVE}{+$3.7\%$} & \textcolor{POSITIVE}{+$9.2\%$}  \\
         \cdottedline{2-14}
         % \cdashline{2-14}[2pt/3pt]\\[-2ex]
         & VCD\cite{vcd} & $58.3\%$ & $44.1\%$ & $55.6\%$ & $42.5\%$ & $55.3\%$ & $42.4\%$ & $60.9\%$ & $48.9\%$ & $57.2\%$ & $46.6\%$ & $56.6\%$ & $47.0\%$  \\
         & \textbf{+ReCo} & \textcolor{POSITIVE}{+$0.9\%$} & \textcolor{POSITIVE}{+$5.7\%$} & \textcolor{POSITIVE}{+$2.0\%$} & \textcolor{POSITIVE}{+$6.3\%$} & \textcolor{POSITIVE}{+$1.7\%$} & \textcolor{POSITIVE}{+$6.1\%$} & \textcolor{POSITIVE}{+$0.1\%$} & \textcolor{NEGATIVE}{-$2.0\%$} & \textcolor{POSITIVE}{+$2.6\%$} & \textcolor{NEGATIVE}{-$0.3\%$} & \textcolor{POSITIVE}{+$1.3\%$} & \textcolor{NEGATIVE}{-$1.1\%$} \\
         \cdottedline{2-14}
         % \cdashline{2-14}[2pt/3pt]\\[-2ex]
         % \cmidrule(lr){2-8}
         & M3ID \cite{m3id} & $57.6\%$ & $41.7\%$ & $55.0\%$ & $40.3\%$ & $54.3\%$ & $39.9\%$ & $59.5\%$ & $38.9\%$ & $56.1\%$ & $37.0\%$ & $56.0\%$ & 36.9\%  \\
         & \textbf{+ReCo} & \textcolor{POSITIVE}{+$0.1\%$} & \textcolor{POSITIVE}{+$7.5\%$} & \textcolor{POSITIVE}{+$1.4\%$} & \textcolor{POSITIVE}{+$8.2\%$} & \textcolor{POSITIVE}{+$1.3\%$} & \textcolor{POSITIVE}{+$8.1\%$} & \textcolor{POSITIVE}{+$1.7\%$} & \textcolor{POSITIVE}{+$2.4\%$} & \textcolor{POSITIVE}{+$4.9\%$} & \textcolor{POSITIVE}{+$4.2\%$} & \textcolor{POSITIVE}{+$2.4\%$} & \textcolor{POSITIVE}{+$2.7\%$}  \\
         \cdottedline{2-14}
         % \cdashline{2-14}[2pt/3pt]\\[-2ex]
         & AvisC\cite{avisc} & $64.6\%$ & $62.9$ & $61.7\%$ & $61.0\%$ & $59.2\%$ & $59.5\%$ & $61.5\%$ & $50.3\%$ & $65.3\%$ & $56.3\%$ & $60.3\%$ & $52.3\%$  \\
         & \textbf{+ReCo} & \textcolor{NEGATIVE}{-$2.1\%$} & \textcolor{NEGATIVE}{-$7.1\%$} & \textcolor{POSITIVE}{+$1.4\%$} & \textcolor{NEGATIVE}{-$5.9\%$} & \textcolor{POSITIVE}{+$1.1\%$} & \textcolor{NEGATIVE}{-$3.8\%$} & \textcolor{POSITIVE}{+$0.6\%$} & \textcolor{POSITIVE}{+$0.7\%$} & \textcolor{NEGATIVE}{-$0.9\%$} & \textcolor{NEGATIVE}{-$2.8\%$} & \textcolor{POSITIVE}{+$0.4\%$} & \textcolor{NEGATIVE}{-$2.0\%$} \\        
         \bottomrule
    \end{NiceTabular}
    \vspace{-13pt}
      
\end{table*}
\setlength{\tabcolsep}{6pt} % Default value: 6pt

\subsubsection{POPE: Experimental evaluations} Since POPE \cite{pope} relies on binary (Yes/No) questions about objects in images, the \textit{fading memory effect} is expected to be less severe. However, it is important to assess ReCo’s performance here and determine if it enhances the results. \Cref{tab:pope} presents Accuracy and F1 scores across three question types (Random, Popular, and Adversarial) for MSCoco \cite{mscoco} and A-OKVQA \cite{aokvqa}. Additional results for LlaVA \cite{llava1} and GQA \cite{gqa} are available in the appendix.%Since the POPE benchmark \cite{pope} is based on binary (Yes/No) questions about objects in the image, we expect that the observed \textit{fading memory effect} will not be that severe. However, it is important to examine whether ReCo is performing satisfactorily here also, and whether it helps improve these results as well.
% In \cref{tab:pope}, we present the Accuracy and the F1 score for all three types of questions (Random, Popular, and Adversarial), for two different datasets (MSCoco \cite{mscoco} and A-OKVQA \cite{aokvqa}) while additional results for LlaVA \cite{llava1} and GQA \cite{gqa} can be found on the appendix. 

{\bf Analysis.} We can observe that ReCo consistently improves the performance across all models, questions, and datasets and, more importantly, it does not overfit the CHAIR benchmark, which is more related to the long-range dependency between the input image and the generated text. For MiniGPT4 specifically, ReCo is {\bf one of the only few} black-box approaches that allows the model to comprehend and answer existence questions correctly, as shown in \cref{fig:pope} and further explained and analyzed in \cref{sec:discriminate_minigpt4}.

\subsubsection{AMBER: Experimental evaluations} AMBER \cite{amber} evaluates both the generative and the discriminative capabilities of a VLM, by asking open-ended questions (e.g., ``Describe the image'') as well as multiple yes/no questions (e.g., ``Is the sky sunny?'', or ``Is there a direct contact between the car and the tree?''). The final metric is  
\begin{equation}
\small
    \text{AMBER} = \frac{1}{2}\bigg(100 - \text{CHAIR} + \text{F1}\bigg)
\end{equation}
where CHAIR is employed for the generative questions and F1 is calculated over all the discriminative questions. In \cref{tab:amber_hallusion} we present the results. 

{\bf Analysis.} ReCo provides a significant performance boost for {\bf all} baselines, demonstrating that, despite the minimal training and modifications we need, it is a valuable add-on for hallucination mitigation in both task families.

\begin{table}[h!]
    \scriptsize
    \centering
    \caption{AMBER \cite{amber} (left) and HallusionBench \cite{hallusionbench} (right) results for InstructBLIP \cite{instructblip}, MiniGPT4 \cite{minigpt4}, and Llava \cite{llava1}. ReCo consistently improves the performance of all models.}
    \label{tab:amber_hallusion}
    \begin{NiceTabular}{l c c c c c c}
         \toprule 
         & \multicolumn{3}{c}{AMBER ($\uparrow$)} & \multicolumn{3}{c}{HallusionBench ($\uparrow$)} \\
         \cmidrule(lr){2-4}\cmidrule(lr){5-7}
         Model(\textcolor{POSITIVE}{+ReCo}) & InstructBLIP & MiniGPT4 & LLaVA & InstructBLIP & MiniGPT4 & LLaVA  \\
        \midrule
         Original & $81.4$ (\textcolor{POSITIVE}{+$3.1$}) & $51.5$ (\textcolor{POSITIVE}{+$32.8$}) & $78.2$ (\textcolor{POSITIVE}{+$2.3$}) & $50.5\%$ (\textcolor{POSITIVE}{+$3.6\%$}) & $46.0\%$ (\textcolor{POSITIVE}{+$3.8\%$}) & $51.9\%$ (\textcolor{POSITIVE}{+$6.9\%$})  \\
         % \cdashline{1-3}[2pt/3pt]\\[-2ex]
         VCD \cite{vcd} & $82.6$ (\textcolor{POSITIVE}{+$3.8$}) & $58.3$ (\textcolor{POSITIVE}{+$23.7$})  & $77.9$ (\textcolor{POSITIVE}{+$2.1$})& $49.8\%$ (\textcolor{POSITIVE}{+$1.0\%$}) & $44.7\%$ (\textcolor{POSITIVE}{+$6.1\%$})  & $49.7\%$ (\textcolor{POSITIVE}{+$8.4\%$})  \\
         % \cdashline{1-3}[2pt/3pt]\\[-2ex]
         M3ID \cite{m3id}  & $83.0$ (\textcolor{POSITIVE}{+$3.5$}) & $49.0$ (\textcolor{POSITIVE}{+$33.1$})  & $77.9$ (\textcolor{POSITIVE}{+$2.9$}) & $49.6\%$ (\textcolor{POSITIVE}{+$1.7\%$}) & $46.3\%$ (\textcolor{POSITIVE}{+$4.9\%$})  & $53.4\%$ (\textcolor{POSITIVE}{+$7.0\%$})  \\
         AvisC \cite{avisc}  & $85.0$ (\textcolor{POSITIVE}{+$2.2$}) & $66.0$ (\textcolor{NEGATIVE}{-$02.6$})  & $79.3$ (\textcolor{POSITIVE}{+$2.1$}) & $47.6\%$ (\textcolor{POSITIVE}{+$3.8\%$}) & $44.0\%$ (\textcolor{POSITIVE}{+$5.4\%$})  & $51.5\%$ (\textcolor{POSITIVE}{+$6.6\%$}) \\
         \bottomrule
    \end{NiceTabular}
    % \vspace{-7pt}
\end{table}

% \begin{table}
%     \footnotesize
%     \centering
%     \begin{tabular}{l cc }
%          \toprule 
%          & \multicolumn{2}{c}{Yes/No answers} \\
%          \cmidrule(lr){2-3}
%          Model & Unmodified & \cellcolor{white}\textbf{+ReCo} \\
%         \midrule
%          MiniGPT4 \cite{minigpt4} & $29.8\%$ & \textcolor{POSITIVE}{$100\%$}   \\
%          VCD \cite{vcd} & $46.6\%$ & \textcolor{POSITIVE}{$100\%$}   \\
%          M3ID \cite{m3id} & $32.1\%$ & \textcolor{POSITIVE}{$100\%$}   \\
%          AvisC \cite{avisc} & $67.1\%$ & \textcolor{POSITIVE}{$100\%$}   \\
%          \bottomrule
%     \end{tabular}
%     \vspace{-7pt}
%     \caption{How many answers to discriminative questions {\em explicitly} contain either ``Yes'' or ``No''? MiniGPT4 out of the box as well as other methods were not able to answer properly discriminative questions in most cases. In contrast, ReCo offers the model the ability to comprehend/answer such questions.}
%     \label{tab:yes_no}
% \end{table}

\subsubsection{HallusionBench: Experimental evaluations} HallusionBench \cite{hallusionbench} evaluates the discriminative capabilities of a VLM by providing it with (modified) images of charts, tables, and maps and asking related questions (see appendix for examples).

{\bf Analysis.} \Cref{tab:amber_hallusion} shows the average accuracy on the HallusionBench. ReCo {\em always} improves the results, irrespective of which underlying VLM and baseline is used. More importantly, in many cases the rest of the baselines fail to improve the unmodified VLM (e.g., all baselines perform worse in InstructBLIP). This underscores the fact that ReCo is more robust across different and diverse questions, like the ones in HallusionBench.

% \begin{table}[h!]
%     \scriptsize
%     \centering
%     \begin{NiceTabular}{l c c c}
%          \toprule 
%          & \multicolumn{3}{c}{Accuracy ($\uparrow$)} \\
%          \cmidrule(lr){2-4}
%          Model & InstructBLIP & MiniGPT4 & LLaVA \\
%         \midrule
%          Original & $50.5\%$ (\textcolor{POSITIVE}{+$3.6\%$}) & $46.0\%$ (\textcolor{POSITIVE}{+$3.8\%$}) & $51.9\%$ (\textcolor{POSITIVE}{+$6.9\%$})  \\
%          % \cdashline{1-3}[2pt/3pt]\\[-2ex]
%          VCD \cite{vcd} & $49.8\%$ (\textcolor{POSITIVE}{+$1.0\%$}) & $44.7\%$ (\textcolor{POSITIVE}{+$6.1\%$})  & $49.7\%$ (\textcolor{POSITIVE}{+$8.4\%$})  \\
%          % \cdashline{1-3}[2pt/3pt]\\[-2ex]
%          M3ID \cite{m3id}  & $49.6\%$ (\textcolor{POSITIVE}{+$1.7\%$}) & $46.3\%$ (\textcolor{POSITIVE}{+$4.9\%$})  & $53.4\%$ (\textcolor{POSITIVE}{+$7.0\%$})  \\
%         % \textbf{+ReCo} & \textcolor{POSITIVE}{+$3.5$} & \textcolor{POSITIVE}{+$33.1$}\\
%         %  \cdashline{1-3}[2pt/3pt]\\[-2ex]
%          AvisC \cite{avisc}  & $47.6\%$ (\textcolor{POSITIVE}{+$3.8\%$}) & $44.0\%$ (\textcolor{POSITIVE}{+$5.4\%$})  & $51.5\%$ (\textcolor{POSITIVE}{+$6.6\%$})  \\
%         % \textbf{+ReCo} & \textcolor{POSITIVE}{+$2.2$} & \textcolor{NEGATIVE}{-$4.6$}\\
%          \bottomrule
%     \end{NiceTabular}
%     \vspace{-7pt}
%     \caption{HallusionBench \cite{amber} results for InstructBLIP \cite{instructblip}, MiniGPT4 \cite{minigpt4}, and Llava \cite{llava1}. ReCo consistently improves the performance of all VLMs by as much as $8.4\%$ (absolute improvement).}
%     \label{tab:hallusion}
% \end{table}

\subsubsection{On the discriminative capabilities of MiniGPT4}\label{sec:discriminate_minigpt4} 

\begin{wrapfigure}{l}{123pt}
\vspace{-13pt}
    \centering
\includegraphics[width=123pt]{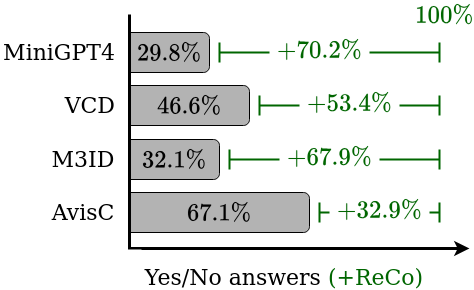}
\vspace{-15pt}
\caption{%How many answers to discriminative questions {\em explicitly} contain either ``Yes'' or ``No''? 
MiniGPT4 out of the box as well as other methods are not able to answer properly (with a \textit{Yes} or \textit{No}) discriminative questions in most cases. In contrast, ReCo offers the model the ability to comprehend/answer such questions.}
\label{tab:yes_no}
\vspace{-10pt}
\end{wrapfigure}

\begin{wrapfigure}{r}{0.44\textwidth}
    \centering
    \vspace{-30pt}
    \includegraphics[width=\linewidth]{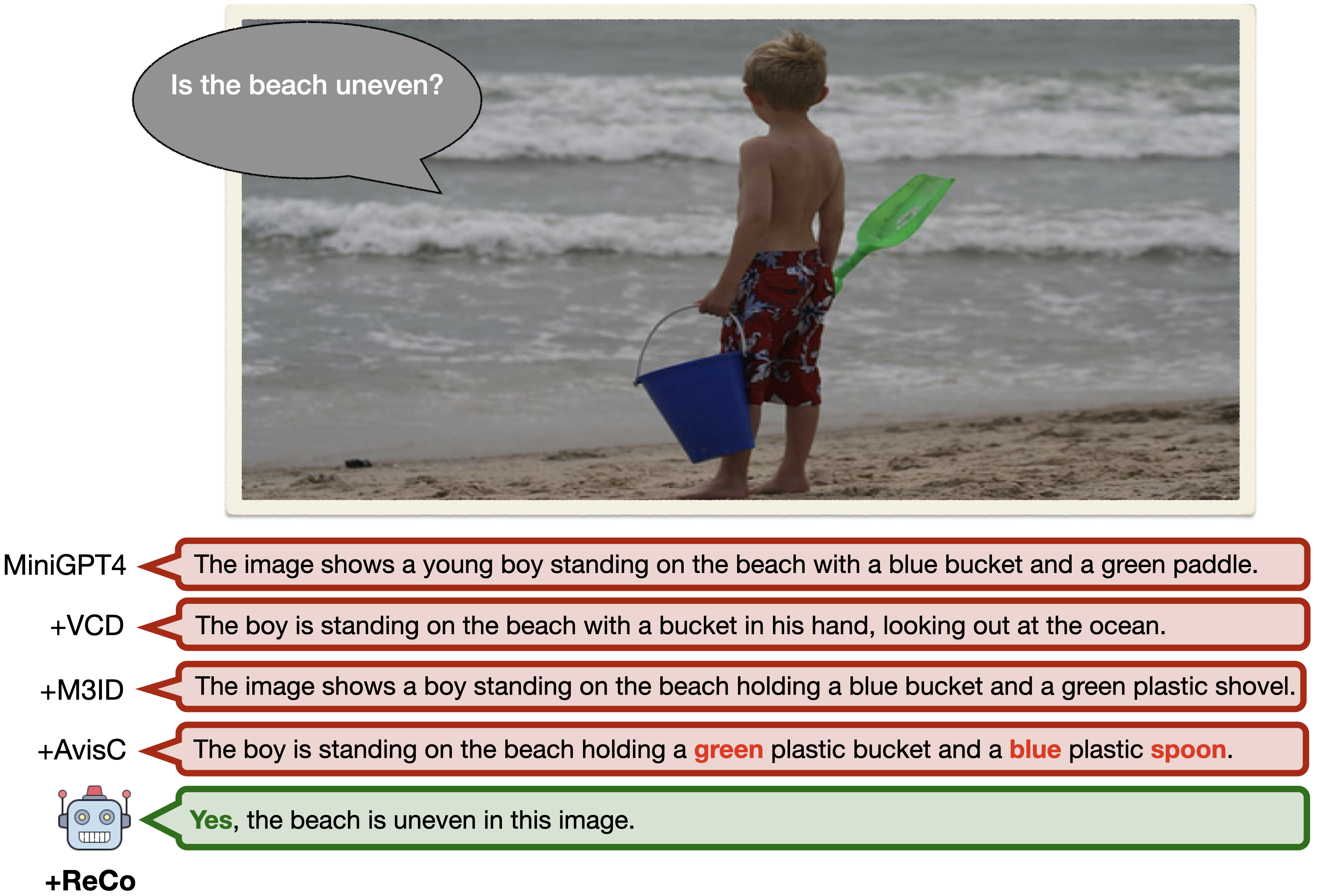}
    \vspace{-17pt}
    \caption{Failure of all MiniGPT4-based models but ReCo to answer the AMBER \cite{amber} questions coherently. All models describe the image (with some of them getting the details wrong) although the question is a binary (yes/no) one (whose true label is ``Yes'').}
    \label{fig:amber}
    \vspace{-5pt}
\end{wrapfigure}

As we see in \cref{fig:pope}, MiniGPT4 is not able to answer existence questions in most of the cases. During AMBER evaluations, we observed that this deficiency extends to all types of discriminative questions. In \cref{tab:yes_no} we show that MiniGPT4, as well as its modifications, do not possess the ability to answer many discriminative (i.e., yes/no) questions, and a failure example is presented in \cref{fig:amber}. As shown, MiniGPT4 is able to understand and answer with either ``Yes'' or ``No'' (independent of the true response label) in less than $30\%$ of the cases, and, while the percentage increases, existing works also face the same difficulty. This, however, means that the results obtained by all baselines on POPE and AMBER do not accurately reflect reality, as these models are not well suited for such questions and small modifications to the evaluation scripts can lead to very different values of accuracy and F1 score. On the contrary, ReCo is able to answer such questions with a $100\%$ rate in all cases, offering a useful new capability to the underlying VLM.

\begin{figure*}[h]
    \centering
    \includegraphics[width=\textwidth]{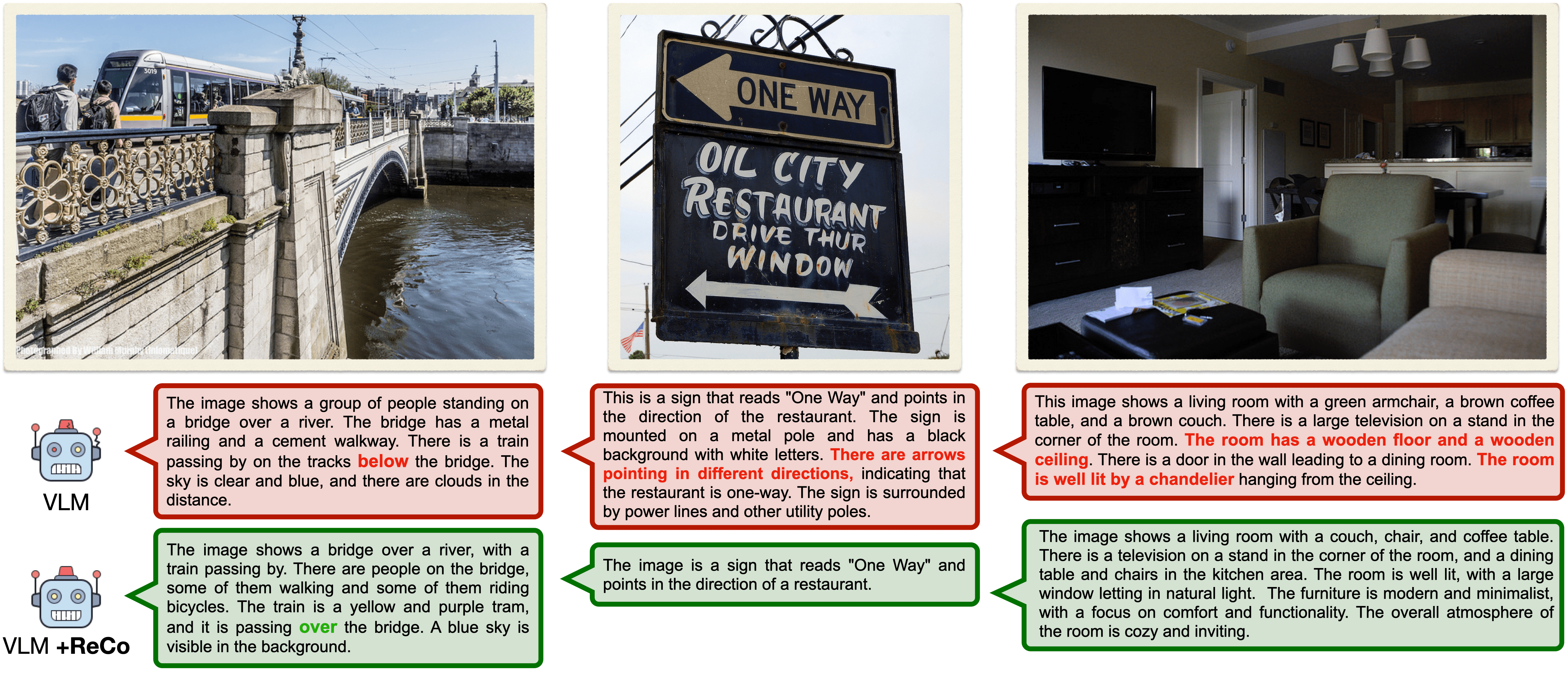}
    \vspace{-16pt}
    \caption{Structural hallucinations: MiniGPT4 \cite{minigpt4} before \textit{(red)} and after ReCo \textit{(green)}. The unmodified VLM tends to get small details about the scene wrong, like the texture of the floor or the written signs that are depicted in the image. Enabling ReCo fixes such mistakes.}
    \label{fig:structure_hallucinations}
\end{figure*}

\subsubsection{Structural hallucinations} CHAIR \cite{chair} and POPE \cite{pope} evaluate only object-related hallucinations. AMBER \cite{amber}
evaluates the model only in the context of yes/no questions although some of its questions are about object relationships and their attributes. However, in our experiments, we observed that beyond a significant reduction in such hallucinations, ReCo is also able to effectively reduce structure-related hallucinations (e.g., relative positions of the objects, object attributes, and so on) in the images' description. In \cref{fig:structure_hallucinations}, we show a few examples where the effect of ReCo is apparent in reducing such hallucinations also. From relative positions to textures and text signs, the improvement of ReCo is apparent in all these cases. Interestingly, we can observe that ReCo does not change the output of the VLM completely, rather it ``intervenes'' only when actually needed, leaving the remainder of the output almost intact. This is of course a by-product of the fact that we treat the VLM as a frozen black-box and we only change the input to the prediction head.

\section{Related Work}\label{sec:works}

\noindent\textbf{Hallucinations mitigation.} Despite their unique capabilities, it is well-known that VLMs hallucinate during the generation process. Multiple works have proposed modified loss functions, optimization schemes, and new datasets \cite{dpo, hadpo, hacl, hallucidoctor} that can improve the performance of the models, albeit extensive re-training of the whole (or a large part) architecture is sometimes needed. A different line of work has proposed modified generation processes: usually, a contrastive decoding approach that ``boosts'' the influence of the visual input to the output \cite{opera, m3id, avisc, vcd}. A detailed review can be found in \cite{surveyhallucination}. Like many of these works, our work also treats the model as a frozen, black box and it intervenes only in the next token prediction. However, similar to the first family of approaches, it is data-driven and it can benefit from the newly proposed datasets and optimization techniques.

\noindent\textbf{Vector Symbolic Architectures.} Ideas describing Vector Symbolic Architectures (VSAs) date back to the 1990s where one of the focus was on introducing operations for efficiently combining (binding) multiple vectors together \cite{hrr, vtb}. The motivation stems from ideas in Symbolic AI where one sought to combine symbols into more complicated sentences but VSAs take also advantage of the representation power of high-dimensional vectors. Multiple works proposed different instantiations of the bind and bundle operations, each one with different performance profiles. A complete review can be found in \cite{vsa}. Recently, some results have infused such ideas into deep learning \cite{fockbox}, achieving a more explicit  compositional behavior, in problems ranging from extreme multi-label classification \cite{hrr_compositions} to a reformulation of the attention mechanism \cite{hrrformer}. 

\section{Conclusions}\label{sec:conclusion}

While VLMs should operate in a way that the next token generation is conditioned on an entity that is a composition of the visual and textual input, this is often 
not the case in practice. Our work describes Reminder Composition (ReCo), a modification to the output of any VLM, which explicitly composes  the visual and textual information. This modification requires minimal training, and despite treating the VLM as a black box is able to significantly improve VLM's forgetting behavior. As a result, we can significantly reduce the hallucination rate of these models. Additionally, ReCo is compatible with other works in hallucination mitigation, and their combination improves further the results across all models and benchmarks.

\textbf{Impact \& Limitations.} ReCo can greatly improve VLMs and help in their smooth and non-harmful usage by all types of users. However, like other mitigation methods, it is not a silver bullet and can benefit from richer compositional rules, access to the model's hidden states, and a deeper integration with the VLM pretraining process. 

%Nevertheless, ReCo proposes a new paradigm by treating the VLM as a black box while, at the same time, benefiting from the improvements in the data-driven approaches.

% \section*{References}

\bibliographystyle{unsrtnat}
\bibliography{main}

\begin{thebibliography}{53}
\providecommand{\natexlab}[1]{#1}
\providecommand{\url}[1]{\texttt{#1}}
\expandafter\ifx\csname urlstyle\endcsname\relax
  \providecommand{\doi}[1]{doi: #1}\else
  \providecommand{\doi}{doi: \begingroup \urlstyle{rm}\Url}\fi

\bibitem[Dai et~al.(2023)Dai, Li, Li, Tiong, Zhao, Wang, Li, Fung, and Hoi]{instructblip}
Wenliang Dai, Junnan Li, Dongxu Li, Anthony Tiong, Junqi Zhao, Weisheng Wang, Boyang Li, Pascale Fung, and Steven Hoi.
\newblock Instruct{BLIP}: Towards general-purpose vision-language models with instruction tuning.
\newblock In \emph{Thirty-seventh Conference on Neural Information Processing Systems}, 2023.
\newblock URL \url{https://openreview.net/forum?id=vvoWPYqZJA}.

\bibitem[Liu et~al.(2023)Liu, Li, Wu, and Lee]{llava1}
Haotian Liu, Chunyuan Li, Qingyang Wu, and Yong~Jae Lee.
\newblock Visual instruction tuning.
\newblock In \emph{Advances in Neural Information Processing Systems}, volume~36. Curran Associates, Inc., 2023.
\newblock URL \url{https://proceedings.neurips.cc/paper_files/paper/2023/file/6dcf277ea32ce3288914faf369fe6de0-Paper-Conference.pdf}.

\bibitem[Chu et~al.(2023)Chu, Xu, Zhou, Yang, Zhang, Yan, Zhou, and Zhou]{qwen_audio}
Yunfei Chu, Jin Xu, Xiaohuan Zhou, Qian Yang, Shiliang Zhang, Zhijie Yan, Chang Zhou, and Jingren Zhou.
\newblock Qwen-audio: Advancing universal audio understanding via unified large-scale audio-language models.
\newblock \emph{arXiv preprint arXiv:2311.07919}, 2023.

\bibitem[Fathullah et~al.(2024)Fathullah, Wu, Lakomkin, Li, Jia, Shangguan, Mahadeokar, Kalinli, Fuegen, and Seltzer]{audiochatllama}
Yassir Fathullah, Chunyang Wu, Egor Lakomkin, Ke~Li, Junteng Jia, Yuan Shangguan, Jay Mahadeokar, Ozlem Kalinli, Christian Fuegen, and Mike Seltzer.
\newblock {A}udio{C}hat{L}lama: Towards general-purpose speech abilities for {LLM}s.
\newblock In \emph{Proceedings of the 2024 Conference of the North American Chapter of the Association for Computational Linguistics: Human Language Technologies (Volume 1: Long Papers)}, Mexico City, Mexico, 2024. Association for Computational Linguistics.
\newblock \doi{10.18653/v1/2024.naacl-long.309}.
\newblock URL \url{https://aclanthology.org/2024.naacl-long.309}.

\bibitem[Zhang et~al.(2023)Zhang, Li, and Bing]{videollama}
Hang Zhang, Xin Li, and Lidong Bing.
\newblock Video-llama: An instruction-tuned audio-visual language model for video understanding.
\newblock \emph{arXiv preprint arXiv:2306.02858}, 2023.
\newblock URL \url{https://arxiv.org/abs/2306.02858}.

\bibitem[Zhu et~al.(2023)Zhu, Chen, Shen, Li, and Elhoseiny]{minigpt4}
Deyao Zhu, Jun Chen, Xiaoqian Shen, Xiang Li, and Mohamed Elhoseiny.
\newblock Minigpt-4: Enhancing vision-language understanding with advanced large language models.
\newblock \emph{arXiv preprint arXiv:2304.10592}, 2023.

\bibitem[Chen et~al.(2023)Chen, Zhu, Shen, Li, Liu, Zhang, Krishnamoorthi, Chandra, Xiong, and Elhoseiny]{minigptv2}
Jun Chen, Deyao Zhu, Xiaoqian Shen, Xiang Li, Zechun Liu, Pengchuan Zhang, Raghuraman Krishnamoorthi, Vikas Chandra, Yunyang Xiong, and Mohamed Elhoseiny.
\newblock Minigpt-v2: large language model as a unified interface for vision-language multi-task learning, 2023.
\newblock URL \url{https://arxiv.org/abs/2310.09478}.

\bibitem[Liu et~al.(2024{\natexlab{a}})Liu, Li, Li, and Lee]{llava2}
Haotian Liu, Chunyuan Li, Yuheng Li, and Yong~Jae Lee.
\newblock Improved baselines with visual instruction tuning.
\newblock In \emph{Proceedings of the IEEE/CVF Conference on Computer Vision and Pattern Recognition (CVPR)}, 2024{\natexlab{a}}.

\bibitem[Touvron et~al.(2023)Touvron, Martin, Stone, Albert, Almahairi, Babaei, Bashlykov, Batra, Bhargava, Bhosale, Bikel, Blecher, Ferrer, Chen, Cucurull, Esiobu, Fernandes, Fu, Fu, Fuller, Gao, Goswami, Goyal, Hartshorn, Hosseini, Hou, Inan, Kardas, Kerkez, Khabsa, Kloumann, Korenev, Koura, Lachaux, Lavril, Lee, Liskovich, Lu, Mao, Martinet, Mihaylov, Mishra, Molybog, Nie, Poulton, Reizenstein, Rungta, Saladi, Schelten, Silva, Smith, Subramanian, Tan, Tang, Taylor, Williams, Kuan, Xu, Yan, Zarov, Zhang, Fan, Kambadur, Narang, Rodriguez, Stojnic, Edunov, and Scialom]{llama}
Hugo Touvron, Louis Martin, Kevin Stone, Peter Albert, Amjad Almahairi, Yasmine Babaei, Nikolay Bashlykov, Soumya Batra, Prajjwal Bhargava, Shruti Bhosale, Dan Bikel, Lukas Blecher, Cristian~Canton Ferrer, Moya Chen, Guillem Cucurull, David Esiobu, Jude Fernandes, Jeremy Fu, Wenyin Fu, Brian Fuller, Cynthia Gao, Vedanuj Goswami, Naman Goyal, Anthony Hartshorn, Saghar Hosseini, Rui Hou, Hakan Inan, Marcin Kardas, Viktor Kerkez, Madian Khabsa, Isabel Kloumann, Artem Korenev, Punit~Singh Koura, Marie-Anne Lachaux, Thibaut Lavril, Jenya Lee, Diana Liskovich, Yinghai Lu, Yuning Mao, Xavier Martinet, Todor Mihaylov, Pushkar Mishra, Igor Molybog, Yixin Nie, Andrew Poulton, Jeremy Reizenstein, Rashi Rungta, Kalyan Saladi, Alan Schelten, Ruan Silva, Eric~Michael Smith, Ranjan Subramanian, Xiaoqing~Ellen Tan, Binh Tang, Ross Taylor, Adina Williams, Jian~Xiang Kuan, Puxin Xu, Zheng Yan, Iliyan Zarov, Yuchen Zhang, Angela Fan, Melanie Kambadur, Sharan Narang, Aurelien Rodriguez, Robert Stojnic, Sergey Edunov, and Thomas
  Scialom.
\newblock Llama 2: Open foundation and fine-tuned chat models, 2023.
\newblock URL \url{https://arxiv.org/abs/2307.09288}.

\bibitem[Favero et~al.(2024)Favero, Zancato, Trager, Choudhary, Perera, Achille, Swaminathan, and Soatto]{m3id}
Alessandro Favero, Luca Zancato, Matthew Trager, Siddharth Choudhary, Pramuditha Perera, Alessandro Achille, Ashwin Swaminathan, and Stefano Soatto.
\newblock Multi-modal hallucination control by visual information grounding.
\newblock In \emph{2024 IEEE/CVF Conference on Computer Vision and Pattern Recognition (CVPR)}, 2024.
\newblock \doi{10.1109/CVPR52733.2024.01356}.

\bibitem[Leng et~al.(2024)Leng, Zhang, Chen, Li, Lu, Miao, and Bing]{vcd}
Sicong Leng, Hang Zhang, Guanzheng Chen, Xin Li, Shijian Lu, Chunyan Miao, and Lidong Bing.
\newblock Mitigating object hallucinations in large vision-language models through visual contrastive decoding.
\newblock In \emph{Proceedings of the IEEE/CVF Conference on Computer Vision and Pattern Recognition (CVPR)}, 2024.

\bibitem[Woo et~al.(2024)Woo, Kim, Jang, Choi, and Kim]{avisc}
Sangmin Woo, Donguk Kim, Jaehyuk Jang, Yubin Choi, and Changick Kim.
\newblock Don't miss the forest for the trees: Attentional vision calibration for large vision language models.
\newblock \emph{arXiv preprint arXiv:2405.17820}, 2024.

\bibitem[Rafailov et~al.(2023)Rafailov, Sharma, Mitchell, Manning, Ermon, and Finn]{dpo}
Rafael Rafailov, Archit Sharma, Eric Mitchell, Christopher~D Manning, Stefano Ermon, and Chelsea Finn.
\newblock Direct preference optimization: Your language model is secretly a reward model.
\newblock In \emph{Thirty-seventh Conference on Neural Information Processing Systems}, 2023.
\newblock URL \url{https://openreview.net/forum?id=HPuSIXJaa9}.

\bibitem[Zhao et~al.(2023)Zhao, Wang, Ouyang, Dong, Wang, and He]{hadpo}
Zhiyuan Zhao, Bin Wang, Linke Ouyang, Xiaoyi Dong, Jiaqi Wang, and Conghui He.
\newblock Beyond hallucinations: Enhancing lvlms through hallucination-aware direct preference optimization, 2023.

\bibitem[Yu et~al.(2024)Yu, Li, Wei, Pang, Ye, Qin, Tang, Tian, and Zhuang]{hallucidoctor}
Qifan Yu, Juncheng Li, Longhui Wei, Liang Pang, Wentao Ye, Bosheng Qin, Siliang Tang, Qi~Tian, and Yueting Zhuang.
\newblock Hallucidoctor: Mitigating hallucinatory toxicity in visual instruction data.
\newblock In \emph{Proceedings of the IEEE/CVF Conference on Computer Vision and Pattern Recognition (CVPR)}, 2024.

\bibitem[Jiang et~al.(2024)Jiang, Xu, Dong, Chen, Ye, Yan, Ye, Zhang, Huang, and Zhang]{hacl}
Chaoya Jiang, Haiyang Xu, Mengfan Dong, Jiaxing Chen, Wei Ye, Ming Yan, Qinghao Ye, Ji~Zhang, Fei Huang, and Shikun Zhang.
\newblock Hallucination augmented contrastive learning for multimodal large language model.
\newblock In \emph{Proceedings of the IEEE/CVF Conference on Computer Vision and Pattern Recognition (CVPR)}, 2024.

\bibitem[Huang et~al.(2024)Huang, Dong, Zhang, Wang, He, Wang, Lin, Zhang, and Yu]{opera}
Qidong Huang, Xiaoyi Dong, Pan Zhang, Bin Wang, Conghui He, Jiaqi Wang, Dahua Lin, Weiming Zhang, and Nenghai Yu.
\newblock Opera: Alleviating hallucination in multi-modal large language models via over-trust penalty and retrospection-allocation.
\newblock In \emph{Proceedings of the IEEE/CVF Conference on Computer Vision and Pattern Recognition}, 2024.

\bibitem[Liu et~al.(2024{\natexlab{b}})Liu, Ye, Xing, and Zou]{vti}
Sheng Liu, Haotian Ye, Lei Xing, and James Zou.
\newblock Reducing hallucinations in vision-language models via latent space steering, 2024{\natexlab{b}}.
\newblock URL \url{https://arxiv.org/abs/2410.15778}.

\bibitem[Nagarajan and Grauman(2018)]{attributes_operators}
Tushar Nagarajan and Kristen Grauman.
\newblock Attributes as operators: Factorizing unseen attribute-object compositions.
\newblock In \emph{Computer Vision -- ECCV 2018}, Cham, 2018. Springer International Publishing.
\newblock ISBN 978-3-030-01246-5.

\bibitem[Aragón-González et~al.(2001)Aragón-González, Aragón, Dávila, Gómez-Rodríguez, and Rodríguez-Andrade]{geometric1}
Gerardo Aragón-González, José Aragón, F.~Dávila, Alfredo Gómez-Rodríguez, and Marco Rodríguez-Andrade.
\newblock \emph{Modern Geometric Calculations in Crystallography}.
\newblock 2001.
\newblock ISBN 0-8176-4199-8.
\newblock \doi{10.1007/978-1-4612-0159-5_18}.

\bibitem[Chisolm(2012)]{geometric2}
Eric Chisolm.
\newblock Geometric algebra, 2012.
\newblock URL \url{https://arxiv.org/abs/1205.5935}.

\bibitem[Naeem et~al.(2021)Naeem, Xian, Tombari, and Akata]{graph_compositional}
MF~Naeem, Y~Xian, F~Tombari, and Zeynep Akata.
\newblock Learning graph embeddings for compositional zero-shot learning.
\newblock In \emph{34th IEEE Conference on Computer Vision and Pattern Recognition}. IEEE, 2021.

\bibitem[Mancini et~al.(2021)Mancini, Naeem, Xian, and Akata]{open_world_compositions}
Massimiliano Mancini, Muhammad~Ferjad Naeem, Yongqin Xian, and Zeynep Akata.
\newblock Open world compositional zero-shot learning.
\newblock In \emph{2021 IEEE/CVF Conference on Computer Vision and Pattern Recognition (CVPR)}, 2021.
\newblock \doi{10.1109/CVPR46437.2021.00518}.

\bibitem[Krishna et~al.(2017)Krishna, Zhu, Groth, Johnson, Hata, Kravitz, Chen, Kalantidis, Li, Shamma, Bernstein, and Fei-Fei]{visual_genome}
Ranjay Krishna, Yuke Zhu, Oliver Groth, Justin Johnson, Kenji Hata, Joshua Kravitz, Stephanie Chen, Yannis Kalantidis, Li-Jia Li, David~A. Shamma, Michael~S. Bernstein, and Li~Fei-Fei.
\newblock Visual genome: Connecting language and vision using crowdsourced dense image annotations.
\newblock \emph{Int. J. Comput. Vision}, 123\penalty0 (1), 2017.
\newblock ISSN 0920-5691.
\newblock \doi{10.1007/s11263-016-0981-7}.
\newblock URL \url{https://doi.org/10.1007/s11263-016-0981-7}.

\bibitem[Misra et~al.(2017)Misra, Gupta, and Hebert]{abhinav_compositions}
Ishan Misra, Abhinav Gupta, and Martial Hebert.
\newblock From red wine to red tomato: Composition with context.
\newblock In \emph{2017 IEEE Conference on Computer Vision and Pattern Recognition (CVPR)}, 2017.
\newblock \doi{10.1109/CVPR.2017.129}.

\bibitem[Yi et~al.(2018)Yi, Wu, Gan, Torralba, Kohli, and Tenenbaum]{neuro_vqa}
Kexin Yi, Jiajun Wu, Chuang Gan, Antonio Torralba, Pushmeet Kohli, and Joshua~B. Tenenbaum.
\newblock Neural-symbolic vqa: disentangling reasoning from vision and language understanding.
\newblock In \emph{Proceedings of the 32nd International Conference on Neural Information Processing Systems}, NIPS'18, Red Hook, NY, USA, 2018. Curran Associates Inc.

\bibitem[Chytas et~al.(2024)Chytas, Kim, and Singh]{multi_compositions}
Sotirios~Panagiotis Chytas, Hyunwoo~J. Kim, and Vikas Singh.
\newblock Understanding multi-compositional learning in vision and language models via category theory.
\newblock In \emph{Computer Vision – ECCV 2024: 18th European Conference, Milan, Italy, 2024}, 2024.

\bibitem[Ganesan et~al.(2021)Ganesan, Gao, Gandhi, Raff, Oates, Holt, and McLean]{hrr_compositions}
Ashwinkumar Ganesan, Hang Gao, Sunil Gandhi, Edward Raff, Tim Oates, James Holt, and Mark McLean.
\newblock Learning with holographic reduced representations.
\newblock In \emph{Advances in Neural Information Processing Systems}, 2021.
\newblock URL \url{https://openreview.net/forum?id=RX6PrcpXP-}.

\bibitem[Gunn and De~Keninck(2019)]{geometric_graphics}
Charles~G. Gunn and Steven De~Keninck.
\newblock Geometric algebra and computer graphics.
\newblock In \emph{ACM SIGGRAPH 2019 Courses}, SIGGRAPH '19, New York, NY, USA, 2019. Association for Computing Machinery.
\newblock ISBN 9781450363075.
\newblock \doi{10.1145/3305366.3328099}.
\newblock URL \url{https://doi.org/10.1145/3305366.3328099}.

\bibitem[Brehmer et~al.(2023)Brehmer, Haan, Behrends, and Cohen]{geometric_transformer}
Johann Brehmer, Pim~De Haan, S{\"o}nke Behrends, and Taco Cohen.
\newblock Geometric algebra transformer.
\newblock In \emph{Thirty-seventh Conference on Neural Information Processing Systems}, 2023.
\newblock URL \url{https://openreview.net/forum?id=M7r2CO4tJC}.

\bibitem[Ruhe et~al.(2023)Ruhe, Gupta, De~Keninck, Welling, and Brandstetter]{clifford_networks}
David Ruhe, Jayesh~K. Gupta, Steven De~Keninck, Max Welling, and Johannes Brandstetter.
\newblock Geometric clifford algebra networks.
\newblock In \emph{Proceedings of the 40th International Conference on Machine Learning}, ICML'23. JMLR.org, 2023.

\bibitem[Li et~al.(2023)Li, Du, Zhou, Wang, Zhao, and Wen]{pope}
Yifan Li, Yifan Du, Kun Zhou, Jinpeng Wang, Xin Zhao, and Ji-Rong Wen.
\newblock Evaluating object hallucination in large vision-language models.
\newblock In \emph{The 2023 Conference on Empirical Methods in Natural Language Processing}, 2023.
\newblock URL \url{https://openreview.net/forum?id=xozJw0kZXF}.

\bibitem[Rohrbach et~al.(2018)Rohrbach, Hendricks, Burns, Darrell, and Saenko]{chair}
Anna Rohrbach, Lisa~Anne Hendricks, Kaylee Burns, Trevor Darrell, and Kate Saenko.
\newblock Object hallucination in image captioning.
\newblock In \emph{Empirical Methods in Natural Language Processing (EMNLP)}, 2018.

\bibitem[Wang et~al.(2023)Wang, Wang, Xu, Zhang, Gu, Jia, Yan, Zhang, and Sang]{amber}
Junyang Wang, Yuhang Wang, Guohai Xu, Jing Zhang, Yukai Gu, Haitao Jia, Ming Yan, Ji~Zhang, and Jitao Sang.
\newblock An llm-free multi-dimensional benchmark for mllms hallucination evaluation.
\newblock \emph{arXiv preprint arXiv:2311.07397}, 2023.

\bibitem[Guan et~al.(2024)Guan, Liu, Wu, Xian, Li, Liu, Wang, Chen, Huang, Yacoob, Manocha, and Zhou]{hallusionbench}
Tianrui Guan, Fuxiao Liu, Xiyang Wu, Ruiqi Xian, Zongxia Li, Xiaoyu Liu, Xijun Wang, Lichang Chen, Furong Huang, Yaser Yacoob, Dinesh Manocha, and Tianyi Zhou.
\newblock Hallusionbench: An advanced diagnostic suite for entangled language hallucination and visual illusion in large vision-language models.
\newblock In \emph{Proceedings of the IEEE/CVF Conference on Computer Vision and Pattern Recognition (CVPR)}, pages 14375--14385, June 2024.

\bibitem[Fu et~al.(2023)Fu, Chen, Shen, Qin, Zhang, Lin, Yang, Zheng, Li, Sun, et~al.]{mme}
Chaoyou Fu, Peixian Chen, Yunhang Shen, Yulei Qin, Mengdan Zhang, Xu~Lin, Jinrui Yang, Xiawu Zheng, Ke~Li, Xing Sun, et~al.
\newblock Mme: A comprehensive evaluation benchmark for multimodal large language models.
\newblock \emph{arXiv preprint arXiv:2306.13394}, 2023.

\bibitem[Wattenberg and Vi{\'e}gas(2024)]{relational}
Martin Wattenberg and Fernanda Vi{\'e}gas.
\newblock Relational composition in neural networks: A survey and call to action.
\newblock In \emph{ICML 2024 Workshop on Mechanistic Interpretability}, 2024.
\newblock URL \url{https://openreview.net/forum?id=zzCEiUIPk9}.

\bibitem[Vaswani et~al.(2017)Vaswani, Shazeer, Parmar, Uszkoreit, Jones, Gomez, Kaiser, and Polosukhin]{attention}
Ashish Vaswani, Noam Shazeer, Niki Parmar, Jakob Uszkoreit, Llion Jones, Aidan~N. Gomez, Lukasz Kaiser, and Illia Polosukhin.
\newblock Attention is all you need.
\newblock 2017.
\newblock URL \url{https://arxiv.org/pdf/1706.03762.pdf}.

\bibitem[Gholamalinezhad and Khosravi(2020)]{pooling}
Hossein Gholamalinezhad and Hossein Khosravi.
\newblock Pooling methods in deep neural networks, a review, 2020.
\newblock URL \url{https://arxiv.org/abs/2009.07485}.

\bibitem[De~Keninck(2024)]{lookmanomatrices}
Steven De~Keninck.
\newblock Look, ma, no matrices!
\newblock In \emph{ACM SIGGRAPH 2024 Talks}, SIGGRAPH '24, New York, NY, USA, 2024. Association for Computing Machinery.
\newblock ISBN 9798400705151.
\newblock \doi{10.1145/3641233.3665801}.
\newblock URL \url{https://doi.org/10.1145/3641233.3665801}.

\bibitem[Plate(1995)]{hrr}
T.A. Plate.
\newblock Holographic reduced representations.
\newblock \emph{IEEE Transactions on Neural Networks}, 6\penalty0 (3), 1995.
\newblock \doi{10.1109/72.377968}.

\bibitem[Gosmann and Eliasmith(2019)]{vtb}
Jan Gosmann and Chris Eliasmith.
\newblock Vector-derived transformation binding: An improved binding operation for deep symbol-like processing in neural networks.
\newblock \emph{Neural Comput.}, 31\penalty0 (5), 2019.
\newblock ISSN 0899-7667.
\newblock \doi{10.1162/neco_a_01179}.
\newblock URL \url{https://doi.org/10.1162/neco_a_01179}.

\bibitem[Thomason(2009)]{logic_ai}
Richmond~H. Thomason.
\newblock Logic and artificial intelligence.
\newblock In \emph{The development of modern logic}. Oxford University Press, 2009.

\bibitem[Fodor and Pylyshyn(1988)]{connectionism}
Jerry~A. Fodor and Zenon~W. Pylyshyn.
\newblock Connectionism and cognitive architecture: A critical analysis.
\newblock \emph{Cognition}, 28\penalty0 (1), 1988.
\newblock ISSN 0010-0277.
\newblock \doi{https://doi.org/10.1016/0010-0277(88)90031-5}.
\newblock URL \url{https://www.sciencedirect.com/science/article/pii/0010027788900315}.

\bibitem[Smolensky(1990)]{ai_bind}
Paul Smolensky.
\newblock Tensor product variable binding and the representation of symbolic structures in connectionist systems.
\newblock \emph{Artificial Intelligence}, 46\penalty0 (1), 1990.
\newblock ISSN 0004-3702.
\newblock \doi{https://doi.org/10.1016/0004-3702(90)90007-M}.
\newblock URL \url{https://www.sciencedirect.com/science/article/pii/000437029090007M}.

\bibitem[Touretzky and Hinton(1985)]{symbols_hinton}
David~S. Touretzky and Geoffrey~E. Hinton.
\newblock Symbols among the neurons: Details of a connectionist inference architecture.
\newblock In \emph{International Joint Conference on Artificial Intelligence}, 1985.
\newblock URL \url{https://api.semanticscholar.org/CorpusID:14138392}.

\bibitem[Schlegel et~al.(2022)Schlegel, Neubert, and Protzel]{vsa}
Kenny Schlegel, Peer Neubert, and Peter Protzel.
\newblock A comparison of vector symbolic architectures.
\newblock \emph{Artif. Intell. Rev.}, 55\penalty0 (6), 2022.
\newblock ISSN 0269-2821.
\newblock \doi{10.1007/s10462-021-10110-3}.
\newblock URL \url{https://doi.org/10.1007/s10462-021-10110-3}.

\bibitem[Wolff et~al.(2018)Wolff, Wirsching, Huber, beim Graben, R{\"o}mer, and Schmitt]{fockbox}
Matthias Wolff, G{\"u}nther Wirsching, Markus Huber, Peter beim Graben, Ronald R{\"o}mer, and Ingo Schmitt.
\newblock A fock space toolbox and some applications in computational cognition.
\newblock In \emph{International Conference on Speech and Computer}, 2018.
\newblock URL \url{https://api.semanticscholar.org/CorpusID:52182870}.

\bibitem[Lin et~al.(2014)Lin, Maire, Belongie, Hays, Perona, Ramanan, Doll{\'a}r, and Zitnick]{mscoco}
Tsung-Yi Lin, Michael Maire, Serge Belongie, James Hays, Pietro Perona, Deva Ramanan, Piotr Doll{\'a}r, and C.~Lawrence Zitnick.
\newblock Microsoft coco: Common objects in context.
\newblock In \emph{Computer Vision -- ECCV 2014}, Cham, 2014. Springer International Publishing.
\newblock ISBN 978-3-319-10602-1.

\bibitem[Schwenk et~al.(2022)Schwenk, Khandelwal, Clark, Marino, and Mottaghi]{aokvqa}
Dustin Schwenk, Apoorv Khandelwal, Christopher Clark, Kenneth Marino, and Roozbeh Mottaghi.
\newblock A-okvqa: A benchmark for visual question answering using world knowledge.
\newblock In \emph{Computer Vision -- ECCV 2022}, Cham, 2022. Springer Nature Switzerland.
\newblock ISBN 978-3-031-20074-8.

\bibitem[Hudson and Manning(2019)]{gqa}
Drew~A Hudson and Christopher~D Manning.
\newblock Gqa: A new dataset for real-world visual reasoning and compositional question answering.
\newblock In \emph{Proceedings of the IEEE/CVF conference on computer vision and pattern recognition}, pages 6700--6709, 2019.

\bibitem[Liu et~al.(2024{\natexlab{c}})Liu, Xue, Chen, Chen, Zhao, Wang, Hou, Li, and Peng]{surveyhallucination}
Hanchao Liu, Wenyuan Xue, Yifei Chen, Dapeng Chen, Xiutian Zhao, Ke~Wang, Liping Hou, Rongjun Li, and Wei Peng.
\newblock A survey on hallucination in large vision-language models, 2024{\natexlab{c}}.
\newblock URL \url{https://arxiv.org/abs/2402.00253}.

\bibitem[Alam et~al.(2023)Alam, Raff, Biderman, Oates, and Holt]{hrrformer}
Mohammad~Mahmudul Alam, Edward Raff, Stella Biderman, Tim Oates, and James Holt.
\newblock Recasting self-attention with holographic reduced representations.
\newblock In \emph{Proceedings of the 40th International Conference on Machine Learning}, volume 202 of \emph{Proceedings of Machine Learning Research}. PMLR, 2023.
\newblock URL \url{https://proceedings.mlr.press/v202/alam23a.html}.

\end{thebibliography}

%%%%%%%%%%%%%%%%%%%%%%%%%%%%%%%%%%%%%%%%%%%%%%%%%%%%%%%%%%%%

\newpage
\appendix
% \clearpage
% \setcounter{page}{1}
% \maketitlesupplementary

\section{ReCo vs Others}

ReCo is uniquely placed in the intersection of methods that treat the underlying VLM as a black box (e.g., \cite{vcd, avisc}) and methods that are training-driven (e.g., \cite{hadpo, vti}). In \cref{tab:comparison}, we show the qualitative advantage of ReCo over multiple other proposed solutions for hallucination mitigation.

\begin{table}[h!]
    \centering
    \setlength{\tabcolsep}{4pt}
    \scriptsize
    \caption{ReCo compared to other models.}
    \label{tab:comparison}
    \begin{tabular}{l|cccc}
        \toprule
         \multirow{2}*{Model} & \multirow{2}*{\shortstack[c]{Training\\ based}} & \multirow{2}*{\shortstack[c]{Black-box\\ VLM}} & \multirow{2}*{\shortstack[c]{Can be deployed\\ during VLM training}}  & \multirow{2}*{\shortstack[c]{Single \\ inference pass}}  \\ 
         & & & & \\
         \midrule
         OPERA\cite{opera}  & \textcolor{red}{\xmark} & \textcolor{red}{\xmark} & \textcolor{red}{\xmark} & \textcolor{red}{\xmark} \\
         VCD\cite{vcd}  & \textcolor{red}{\xmark} & \cmark & \textcolor{red}{\xmark} & \textcolor{red}{\xmark} \\
         AvisC\cite{avisc}  & \textcolor{red}{\xmark} & \cmark & \textcolor{red}{\xmark} & \textcolor{red}{\xmark} \\
         M3ID\cite{m3id}  & \textcolor{red}{\xmark} & \cmark & \textcolor{red}{\xmark} & \textcolor{red}{\xmark} \\
         HALC\cite{hacl} & \textcolor{red}{\xmark} & \cmark & \textcolor{red}{\xmark} & \textcolor{red}{\xmark} \\
         VTI\cite{vti}  & \cmark & \textcolor{red}{\xmark} & \textcolor{red}{\xmark} & \cmark \\
         HA-DPO\cite{hadpo}  & \cmark & \textcolor{red}{\xmark} & \cmark & \cmark \\
         \rowcolor{GREENBACK}\textbf{ReCo} & \cmark & \cmark & \cmark & \cmark \\
         \bottomrule
    \end{tabular}
    
\end{table}

\section{Training data}

\begin{figure}[h]
    \centering
    \includegraphics[width=0.4\linewidth]{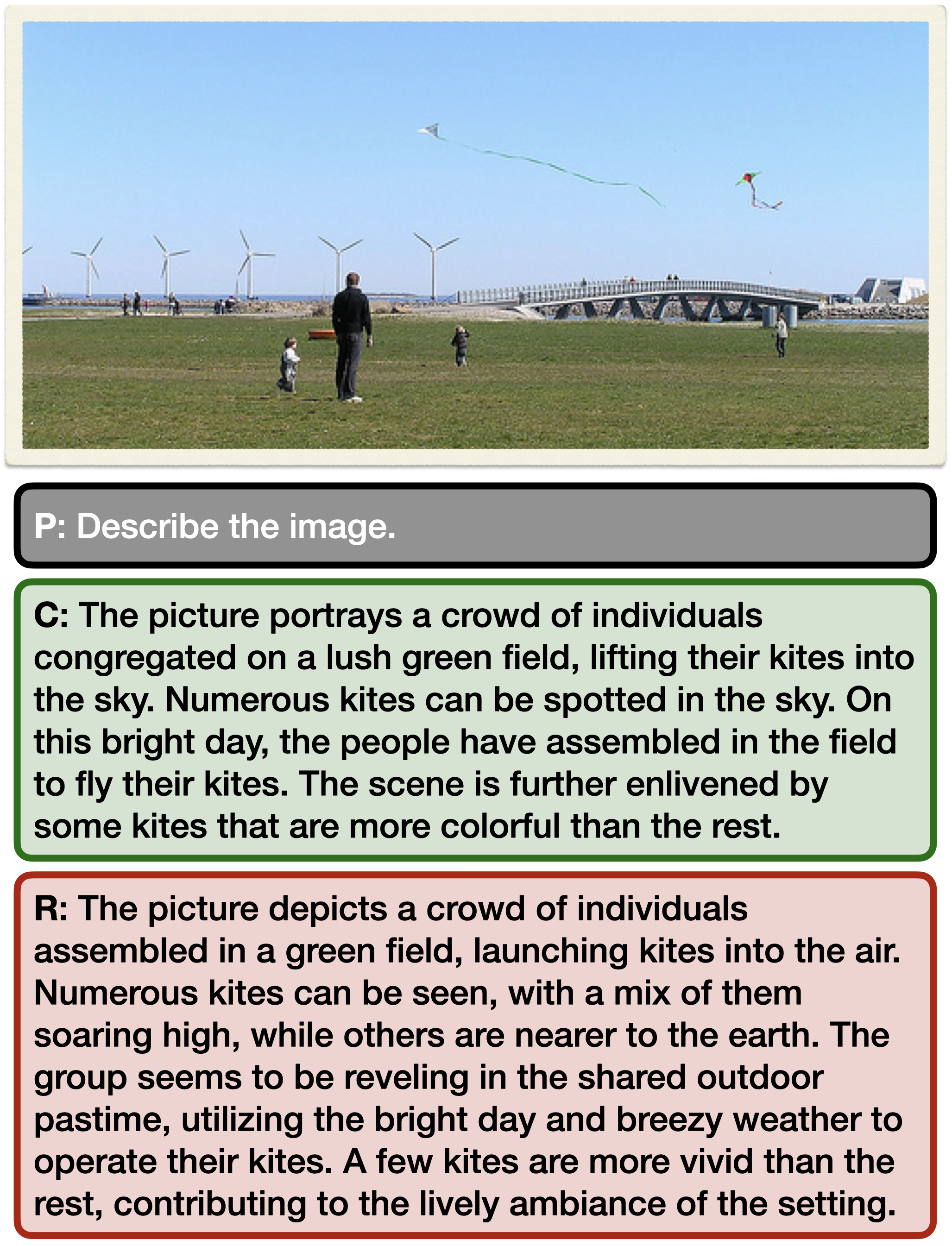}
    \caption{A sample of the HA-DPO dataset \cite{hadpo}. While the $C$ and $R$ prompts may not be entirely accurate since they were generated with the help of LLMs \cite{hadpo}, the Direct Preference Optimization \cite{dpo} successfully trains ReCo.}
    \label{fig:hadpo}
\end{figure}

We train ReCo using HA-DPO \cite{hadpo}. The dataset consists of quadruples $(I, P, C, R)$ where $I$ corresponds to the image and $P$ corresponds to its accompanying prompt (or question). Finally, $C$ and $R$ correspond to \textit{Chosen} and \textit{Rejected} respectively: these are the two answers used for contrastive loss (or Direct Preference Optimization \cite{dpo}). An example can be seen in \cref{fig:hadpo}. The whole dataset consists of only $1853$ images which were extracted from the Visual Genome database \cite{visual_genome}. 

\section{Hyperparameters}

We combine ReCo with two widely used VLMs: InstructBLIP \cite{instructblip} and MiniGPT4 \cite{minigpt4} and we follow the same training procedure for both. The learning rate is set to $5e\text{-}3$ and we train the model for $10$ epochs. Additionally, the $\beta$ and $\lambda$ parameters of HA-DPO are set to $0.8, 0.2$ respectively. Finally, we chose a batch size of $128$.

\section{Benchmarks}

We consider three benchmarks designed to evaluate the hallucinating performance of VLMs across different tasks.

\begin{figure}
    \centering
    \includegraphics[width=0.5\linewidth]{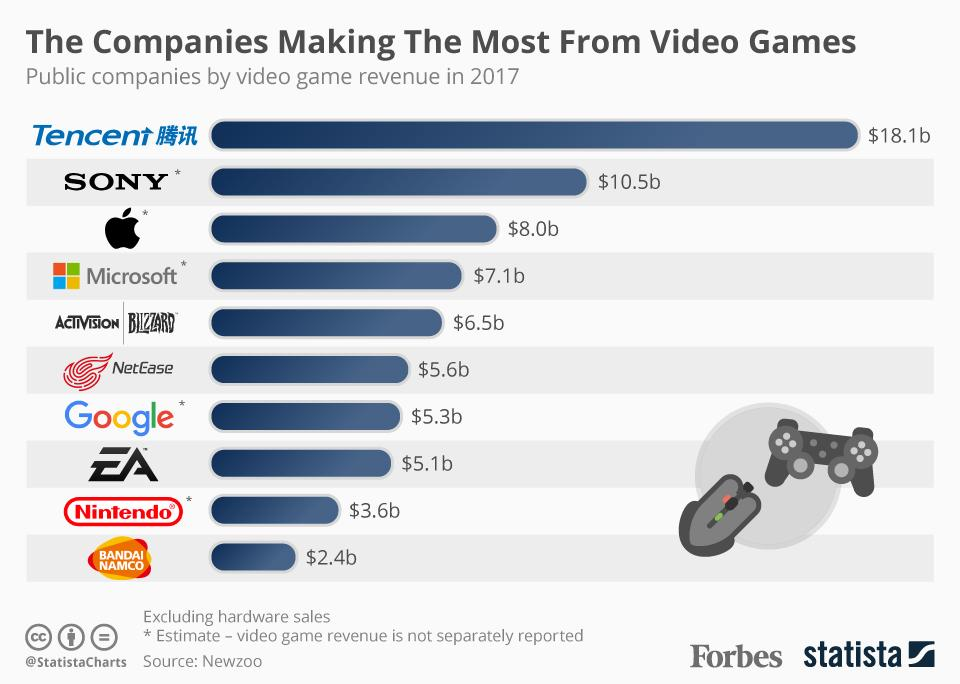}
    \caption{An image obtain from HallusionBench \cite{hallusionbench}. The corresponding question is: ``According to the image, in 2017, was Tencent the company with the highest revenue from video games, with Sony as the second-highest earner?''}
    \label{fig:hallusionbench}
\end{figure}

\begin{enumerate}
    \item \textbf{POPE} \cite{pope}: The POPE benchmark is focused on binary existence questions. It consists of $9000$ questions in total, of the form \textit{``Is there a $\langle$object$\rangle$ in the image?''}, each one accompanied by an image. POPE is designed to measure the disriminative capabilities of the underlying VLM, in the context of existence yes/no questions. The objects that inform the questions are chosen from the whole universe of the depicted objects in three different ways: 
    \begin{itemize}
        \item \textit{Random}: randomly chosen from any of the existing objects in the dataset.
        \item \textit{Popular}: the objects are chosen from the top-k most frequent objects.
        \item \textit{Adversarial}: the objects are chosen based on the co-occuring frequencies with the objects depicted on each image.
    \end{itemize}
    As expected, the ``difficulty'' of the benchmark increases as we transition from \textit{Random} to \textit{Adversarial}, something reflected in our results as well as the existing works too.
    \item \textbf{CHAIR} \cite{chair}: While POPE is focused on the discriminative capabilities of the VLMs, CHAIR assesses their generative power. This benchmark consists of images accompanied by their ground truth labels, i.e., which objects exist in each one of them. The VLM is prompted with a prompt such as \textit{``Describe the image.''}, or \textit{``Provide a detailed description to the image.''}. In our experiments we used \textit{``Describe the image.''} but any such prompt can be used with no changes. After each generated description is obtained, CHAIR estimates the hallucination rate of the VLMs, reporting two metrics:
    \begin{itemize}
        \item \textit{CHAIR$_{\text{i}}$}: CHAIR$_{\text{i}}$ is define as the ratio of the objects in the description that do not really exist in the image, i.e.,
        \begin{equation}
        \small
            \text{CHAIR}_{i} = \frac{|\{\text{hallucinated objects}\}|}{|\{\text{all mentioned objects}\}|}
        \end{equation}
        \item \textit{CHAIR$_{\text{s}}$}: On the contrary, CHAIR$_{\text{s}}$ estimates how much the model ``talks'' about hallucinated objects, and it is defined as:
        \begin{equation}
        \footnotesize
            \text{CHAIR}_{s} = \frac{|\{\text{sentences with hallucinated objects}\}|}{|\{\text{all sentences}\}|}
        \end{equation}
    \end{itemize}
    \item \textbf{AMBER} \cite{amber}: AMBER combines and extends CHAIR and POPE, by evaluating the underlying VLM on both a wide array of discriminative questions about objects relationships (e.g., \textit{``Is there a direct contect between $\langle$object$_1\rangle$ and $\langle$object$_2\rangle$''}), objects attributes (e.g., \textit{``Is the cloud black in this image?''}), and object existence (\textit{``Is there a $\langle$object$\rangle$ in the image?''}), as well as generative questions (\textit{``Describe the image.''}). The AMBER metric is calculated as:
    \begin{equation}
    \small
    \text{AMBER} = \frac{1}{2}\bigg(100 - \text{CHAIR} + \text{F1}\bigg)
    \end{equation}
    with CHAIR$_i$ employed for the generative questions and F1 for the discriminative questions.
    \item \textbf{MME} \cite{mme}: MME is yet another, larger, discriminative benchmark. Similarly to AMBER, MME also consists of multiple types of discriminative questions such as \textit{artwork} related, \textit{code} related, and \textit{numerical} related questions. For each image, MME proposes two questions (one whose correct answer is \textit{`Yes'} and one \textit{`No'}). The total score for each type of questions is the summation of accuracy and accuracy$+$, with accuracy$+$ being the ratio of images in which both corresponding questions were answered correctly. Finally, the total perception score is the summation of all the corresponding scores for each question type. 
    \item \textbf{HallusionBench} \cite{hallusionbench}: Similar to MME, HallusionBench introduces a wide array of different discrimination questions. Differently than the other benchmarks, most of the questions in this benchmark are about charts, tables, and maps depicted on the given images. An example of a image-question pair can be seen in \cref{fig:hallusionbench}. In many cases, the depicted diagrams do not accurately reflect reality and they were twisted in order to evaluate whether a VLM is actually ``looking'' at the image or its answer is based solely on the backbone LLM. Although the questions are discriminative, the long, detailed questions may lead to a more severe \textit{fading memory effect}, compared to other discriminative benchmarks like POPE.
\end{enumerate}

\section{More quantitative results}

\subsection{POPE}

Besides \cref{tab:pope} in which we present the POPE results for InstructBLIP \cite{instructblip} and MiniGPT4 \cite{minigpt4} on MSCoco and A-OKVQA \cite{mscoco, aokvqa}, in \cref{tab:pope_llava} you can observe the results on the same datasets for Llava \cite{llava1} too. The observations from the main text hold here too, with ReCo being a valuable addition to all the considered baselines, improving the performance by as much as $6\%$ (absolute improvement). Finally, in \cref{tab:pope_gqa} you can observe the results for all three VLMs in yet another dataset: GQA \cite{gqa}. Similarly to the previous observations, ReCo consistently improves the results across the board in this dataset too.

\setlength{\tabcolsep}{4pt} % Default value: 6pt

\begin{table*}
    \scriptsize
    \centering
    \captionof{table}{POPE \cite{pope} results on MSCoco \cite{mscoco} and A-OKVQA \cite{aokvqa}, for Llava \cite{llava1}. In both datasets, ReCo provides a significant performance boost to all the methods (where \textit{Original} stands for the unmodified Llava).}
      \label{tab:pope_llava}
    \begin{NiceTabular}{cl|cccccc|cccccc}
         \toprule
         & & \multicolumn{6}{c|}{MSCoco \cite{mscoco}} & \multicolumn{6}{c}{A-OKVQA \cite{aokvqa}} \\
         \cmidrule(lr){3-8}\cmidrule(lr){9-14}
         \multicolumn{2}{l|}{\multirow{1}{*}{Model}} & \multicolumn{2}{c}{Random} & \multicolumn{2}{c}{Popular}  & \multicolumn{2}{c|}{Adversarial} & \multicolumn{2}{c}{Random} & \multicolumn{2}{c}{Popular}  & \multicolumn{2}{c}{Adversarial} \\
          \cmidrule(lr){3-4}\cmidrule(lr){5-6}\cmidrule(lr){7-8}\cmidrule(lr){9-10}\cmidrule(lr){11-12}\cmidrule(lr){13-14}
          &  & Acc ($\uparrow$) & F1 ($\uparrow$) & Acc ($\uparrow$) & F1 ($\uparrow$)  & Acc ($\uparrow$) & F1 ($\uparrow$) & Acc ($\uparrow$) & F1 ($\uparrow$) &Acc ($\uparrow$) & F1 ($\uparrow$)  & Acc ($\uparrow$) & F1 ($\uparrow$) \\
         % \hline 
         % \hline\\[-2ex]
         \midrule
         \multirow{8}*{\rotatebox[origin=c]{90}{Llava \cite{llava1}}}& Original & $84.7\%$ & $85.0\%$ & $81.8\%$ & $82.6\%$ & $76.7\%$ & $78.7\%$ & $81.9\%$ & $83.5\%$ & $75.9\%$ & $79.2\%$ & $68.4\%$ & $74.5\%$ \\
         & \textbf{+ReCo} & \textcolor{POSITIVE}{+$0.5\%$} & \textcolor{NEGATIVE}{-$0.3\%$} & \textcolor{POSITIVE}{+$1.3\%$} & \textcolor{POSITIVE}{+$0.2\%$} & \textcolor{POSITIVE}{+$2.5\%$} & \textcolor{POSITIVE}{+$1.1\%$} & \textcolor{POSITIVE}{+$2.4\%$} & \textcolor{POSITIVE}{+$1.3\%$} & \textcolor{POSITIVE}{+$3.0\%$} & \textcolor{POSITIVE}{+$1.5\%$} & \textcolor{POSITIVE}{+$2.6\%$} & \textcolor{POSITIVE}{+$0.9\%$}  \\
         \cdottedline{2-14}
         % \cdashline{2-14}[2pt/3pt]\\[-2ex]
         & VCD\cite{vcd} & $85.1\%$ & $85.4\%$ & $81.6\%$ & $82.6\%$ & $76.2\%$ & $78.6\%$ & $81.8\%$ & $83.6\%$ & $75.5\%$ & $79.1\%$ & $67.8\%$ & $74.3\%$  \\
         & \textbf{+ReCo} & \textcolor{POSITIVE}{+$2.4\%$} & \textcolor{POSITIVE}{+$1.5\%$} & \textcolor{POSITIVE}{+$2.3\%$} & \textcolor{POSITIVE}{+$1.2\%$} & \textcolor{POSITIVE}{+$3.3\%$} & \textcolor{POSITIVE}{+$1.6\%$} & \textcolor{POSITIVE}{+$3.5\%$} & \textcolor{POSITIVE}{+$2.3\%$} & \textcolor{POSITIVE}{+$5.1\%$} & \textcolor{POSITIVE}{+$3.2\%$} & \textcolor{POSITIVE}{+$4.3\%$} & \textcolor{POSITIVE}{+$2.0\%$}  \\
         \cdottedline{2-14}
         % \cdashline{2-14}[2pt/3pt]\\[-2ex]
         % \cmidrule(lr){2-8}
         & M3ID\cite{m3id} & $86.4\%$ & $86.5\%$ & $82.8\%$ & $83.5\%$ & $77.3\%$ & $79.3\%$ & $83.0\%$ & $84.6\%$ & $76.7\%$ & $79.9\%$ & $68.6\%$ & $74.7\%$  \\
         & \textbf{+ReCo} & \textcolor{POSITIVE}{+$0.7\%$} & \textcolor{NEGATIVE}{-$0.3\%$} & \textcolor{POSITIVE}{+$2.4\%$} & \textcolor{POSITIVE}{+$1.1\%$} & \textcolor{POSITIVE}{+$3.5\%$} & \textcolor{POSITIVE}{+$1.4\%$} & \textcolor{POSITIVE}{+$4.5\%$} & \textcolor{POSITIVE}{+$3.1\%$} & \textcolor{POSITIVE}{+$6.0\%$} & \textcolor{POSITIVE}{+$3.8\%$} & \textcolor{POSITIVE}{+$6.0\%$} & \textcolor{POSITIVE}{+$2.9\%$}  \\
         \cdottedline{2-14}
         % \cdashline{2-14}[2pt/3pt]\\[-2ex]
         & AvisC\cite{avisc} & $87.8\%$ & $87.8\%$ & $83.9\%$ & $84.5\%$ & $78.2\%$ & $80.1\%$ & $84.5\%$ & $85.8\%$ & $78.6\%$ & $81.4\%$ & $69.0\%$ & $75.2\%$  \\
         & \textbf{+ReCo} & \textcolor{POSITIVE}{+$0.8\%$} & \textcolor{POSITIVE}{+$0.2\%$} & \textcolor{POSITIVE}{+$2.9\%$} & \textcolor{POSITIVE}{+$2.0\%$} & \textcolor{POSITIVE}{+$3.5\%$} & \textcolor{POSITIVE}{+$2.0\%$} & \textcolor{POSITIVE}{+$3.4\%$} & \textcolor{POSITIVE}{+$2.4\%$} & \textcolor{POSITIVE}{+$2.9\%$} & \textcolor{POSITIVE}{+$1.6\%$} & \textcolor{POSITIVE}{+$4.0\%$} & \textcolor{POSITIVE}{+$1.9\%$}  \\
         % \cdashline{2-8}[2pt/3pt]\\[-2ex]
         % & HA-DPO$_\text{t}$\cite{hadpo} & $\mathbf{87.9\%}$ & $\mathbf{87.6\%}$ & $\mathbf{84.1\%}$ & $\mathbf{84.7\%}$ & $80.0\%$ & $79.9\%$  \\
         %  & \rowcolor{GREENBACK}\textbf{ReCo}$_\text{t}$ & $0\%$ & $0\%$ & $0\%$ & $0\%$ & $0\%$ & $0\%$  \\
         \bottomrule
    \end{NiceTabular}

\end{table*}

\setlength{\tabcolsep}{6pt} % Default value: 6pt

\begin{table*}
    \scriptsize
    \centering
    \captionof{table}{POPE \cite{pope} results on GQA \cite{gqa}, for all three VLMs (InstructBLIP \cite{instructblip}, MiniGPT4 \cite{minigpt4}, and Llava \cite{llava1}).}
      \label{tab:pope_gqa}
    \begin{NiceTabular}{cl|cccccc}
         \toprule
         & & \multicolumn{6}{c}{GQA \cite{gqa}}  \\
         \cmidrule(lr){3-8}
         \multicolumn{2}{l|}{\multirow{1}{*}{Model}} & \multicolumn{2}{c}{Random} & \multicolumn{2}{c}{Popular}  & \multicolumn{2}{c}{Adversarial} \\
          \cmidrule(lr){3-4}\cmidrule(lr){5-6}\cmidrule(lr){7-8}
          &  & Acc ($\uparrow$) & F1 ($\uparrow$) & Acc ($\uparrow$) & F1 ($\uparrow$)  & Acc ($\uparrow$) & F1 ($\uparrow$) \\
         % \hline 
         % \hline\\[-2ex]
         \midrule
         \multirow{8}*{\rotatebox[origin=c]{90}{InstructBLIP \cite{instructblip}}}& Original & $79.2\%$ & $80.7\%$ & $73.2\%$ & $76.6\%$ & $68.8\%$ & $73.5\%$\\
         & \textbf{+ReCo} & \textcolor{POSITIVE}{+$3.4\%$} & \textcolor{POSITIVE}{+$0.1\%$} & \textcolor{POSITIVE}{+$4.7\%$} & \textcolor{POSITIVE}{+$0.1\%$} & \textcolor{POSITIVE}{+$7.0\%$} & \textcolor{POSITIVE}{+$1.5\%$}  \\
         \cdottedline{2-8}
         % \cdashline{2-8}[2pt/3pt]\\[-2ex]
         & VCD\cite{vcd} & $80.9\%$ & $81.8\%$ & $73.3\%$ & $76.3\%$ & $69.6\%$ & $74.1\%$ \\
         & \textbf{+ReCo} & \textcolor{POSITIVE}{+$2.9\%$} & \textcolor{POSITIVE}{+$0.1\%$} & \textcolor{POSITIVE}{+$6.3\%$} & \textcolor{POSITIVE}{+$2.1\%$} & \textcolor{POSITIVE}{+$6.2\%$} & \textcolor{POSITIVE}{+$1.0\%$}   \\
         \cdottedline{2-8}
         % \cdashline{2-8}[2pt/3pt]\\[-2ex]
         % \cmidrule(lr){2-8}
         & M3ID\cite{m3id} & $81.0\%$ & $82.2\%$ & $74.4\%$ & $77.4\%$ & $69.3\%$ & $74.0\%$   \\
         & \textbf{+ReCo} & \textcolor{POSITIVE}{+$3.4\%$} & \textcolor{POSITIVE}{+$0.1\%$} & \textcolor{POSITIVE}{+$5.1\%$} & \textcolor{POSITIVE}{+$0.6\%$} & \textcolor{POSITIVE}{+$7.1\%$} & \textcolor{POSITIVE}{+$1.3\%$}   \\
         \cdottedline{2-8}
         % \cdashline{2-8}[2pt/3pt]\\[-2ex]
         & AvisC\cite{avisc} & $85.1\%$ & $85.3\%$ & $76.2\%$ & $78.5\%$ & $72.0\%$ & $76.0\%$  \\
         & \textbf{+ReCo} & \textcolor{POSITIVE}{+$0.4\%$} & \textcolor{NEGATIVE}{-$0.3\%$} & \textcolor{POSITIVE}{+$4.0\%$} & \textcolor{POSITIVE}{+$0.8\%$} & \textcolor{POSITIVE}{+$5.1\%$}  & \textcolor{POSITIVE}{+$0.9\%$}  \\
         % \cdashline{2-8}[2pt/3pt]\\[-2ex]
         % & HA-DPO$_\text{t}$\cite{hadpo} & $\mathbf{87.9\%}$ & $\mathbf{87.6\%}$ & $\mathbf{84.1\%}$ & $\mathbf{84.7\%}$ & $80.0\%$ & $79.9\%$  \\
         %  & \rowcolor{GREENBACK}\textbf{ReCo}$_\text{t}$ & $0\%$ & $0\%$ & $0\%$ & $0\%$ & $0\%$ & $0\%$  \\
        \midrule
         \multirow{8}*{\rotatebox[origin=c]{90}{MiniGPT4 \cite{minigpt4}}}& Original & $53.9\%$ & $32.9\%$ & $52.6\%$ & $32.3\%$ & $51.7\%$ & $31.5\%$\\
         & \textbf{+ReCo} & \textcolor{POSITIVE}{+$3.0\%$} & \textcolor{POSITIVE}{+$30.0\%$} & \textcolor{POSITIVE}{+$2.7\%$} & \textcolor{POSITIVE}{+$18.7\%$} & \textcolor{POSITIVE}{+$2.9\%$} & \textcolor{POSITIVE}{+$19.8\%$}  \\
         \cdottedline{2-8}
         % \cdashline{2-8}[2pt/3pt]\\[-2ex]
         & VCD\cite{vcd} & $55.7\%$ & $38.8\%$ & $54.1\%$ & $38.0\%$ & $53.4\%$ & $37.7\%$ \\
         & \textbf{+ReCo} & \textcolor{POSITIVE}{+$0.9\%$} & \textcolor{POSITIVE}{+$11.8\%$} & \textcolor{POSITIVE}{+$0.9\%$} & \textcolor{POSITIVE}{+$11.8\%$} & \textcolor{POSITIVE}{+$0.1\%$} & \textcolor{POSITIVE}{+$11.7\%$}   \\
         \cdottedline{2-8}
         % \cdashline{2-8}[2pt/3pt]\\[-2ex]
         % \cmidrule(lr){2-8}
         & M3ID\cite{m3id} & $55.3\%$ & $37.9\%$ & $54.0\%$ & $37.2\%$ & $52.5\%$ & $35.7\%$   \\
         & \textbf{+ReCo} & \textcolor{POSITIVE}{+$1.5\%$} & \textcolor{POSITIVE}{+$15.5\%$} & \textcolor{POSITIVE}{+$1.4\%$} & \textcolor{POSITIVE}{+$15.4\%$} & \textcolor{POSITIVE}{+$1.6\%$} & \textcolor{POSITIVE}{+$15.8\%$}   \\
         \cdottedline{2-8}
         % \cdashline{2-8}[2pt/3pt]\\[-2ex]
         & AvisC\cite{avisc} & $63.1\%$ & $63.5\%$ & $60.0\%$ & $61.6\%$ & $57.0\%$ & $60.0\%$  \\
         & \textbf{+ReCo} & \textcolor{NEGATIVE}{-$2.4\%$} & \textcolor{NEGATIVE}{-$13.5\%$} & \textcolor{POSITIVE}{+$0.7\%$} & \textcolor{NEGATIVE}{-$12.0\%$} & \textcolor{POSITIVE}{+$0.2\%$}  & \textcolor{NEGATIVE}{-$14.1\%$}  \\ 
         \midrule
         \multirow{8}*{\rotatebox[origin=c]{90}{LLaVA \cite{llava1}}}& Original & $82.3\%$ & $84.0\%$ & $73.9\%$ & $78.1\%$ & $68.6\%$ & $74.8\%$\\
         & \textbf{+ReCo} & \textcolor{POSITIVE}{+$2.5\%$} & \textcolor{POSITIVE}{+$1.6\%$} & \textcolor{POSITIVE}{+$2.3\%$} & \textcolor{POSITIVE}{+$0.9\%$} & \textcolor{POSITIVE}{+$4.2\%$} & \textcolor{POSITIVE}{+$1.9\%$}  \\
         \cdottedline{2-8}
         % \cdashline{2-8}[2pt/3pt]\\[-2ex]
         & VCD\cite{vcd} & $81.9\%$ & $84.0\%$ & $72.5\%$ & $77.5\%$ & $68.1\%$ & $74.9\%$ \\
         & \textbf{+ReCo} & \textcolor{POSITIVE}{+$4.7\%$} & \textcolor{POSITIVE}{+$3.2\%$} & \textcolor{POSITIVE}{+$3.7\%$} & \textcolor{POSITIVE}{+$1.8\%$} & \textcolor{POSITIVE}{+$4.5\%$} & \textcolor{POSITIVE}{+$1.9\%$}   \\
         \cdottedline{2-8}
         % \cdashline{2-8}[2pt/3pt]\\[-2ex]
         % \cmidrule(lr){2-8}
         & M3ID\cite{m3id} & $83.6\%$ & $85.3\%$ & $74.3\%$ & $78.7\%$ & $68.8\%$ & $75.1\%$   \\
         & \textbf{+ReCo} & \textcolor{POSITIVE}{+$3.8\%$} & \textcolor{POSITIVE}{+$2.3\%$} & \textcolor{POSITIVE}{+$5.5\%$} & \textcolor{POSITIVE}{+$2.8\%$} & \textcolor{POSITIVE}{+$6.6\%$} & \textcolor{POSITIVE}{+$3.3\%$}   \\
         \cdottedline{2-8}
         % \cdashline{2-8}[2pt/3pt]\\[-2ex]
         & AvisC\cite{avisc} & $84.6\%$ & $86.1\%$ & $74.6\%$ & $79.0\%$ & $69.4\%$ & $75.8\%$  \\
         & \textbf{+ReCo} & \textcolor{POSITIVE}{+$3.4\%$} & \textcolor{POSITIVE}{+$2.3\%$} & \textcolor{POSITIVE}{+$5.1\%$} & \textcolor{POSITIVE}{+$3.0\%$} & \textcolor{POSITIVE}{+$6.0\%$}  & \textcolor{POSITIVE}{+$3.2\%$}  \\ 
         \bottomrule
    \end{NiceTabular}

\end{table*}

\subsection{CHAIR}

In \cref{tab:chair_llava}, we depict the improvement that ReCo offers over all baselines in the generative task of image description, when the underlying VLM is LlaVA \cite{llava1}. Similar to the observations of the main text, ReCo reduces both CHAIR$_s$ and CHAIR$_i$, especially as the output length increases, with the improvement being as high as $44\%$ and $33\%$, respectively.

\setlength{\tabcolsep}{4.5pt} % Default value: 6pt

\begin{table*}[h]
    \scriptsize
    \centering
    \caption{CHAIR \cite{chair} results for Llava \cite{llava1}. \textit{Original} stands for the unmodified VLM, while VCD \cite{vcd}, M3ID \cite{m3id}, and AvisC \cite{avisc} are the three baselines we consider. In all four cases, the addition of ReCo improves significantly the performance as the generation progresses, reducing CHAIR$_\text{s}$ as much as $44\%$ and CHAIR$_\text{i}$ $33\%$, with $32,\ 64,\ 128,\ 256,\ 512,\ \text{and } 1024$ standing for the maximum allowed number of generated tokens.}
    \label{tab:chair_llava}
    \begin{NiceTabular}{cl|c c c c c c | c c c c c c | c }
         \toprule
          \multicolumn{2}{l|}{\multirow{2}{*}{Model}} & \multicolumn{6}{c}{CHAIR$_\text{s}(\downarrow)$} & \multicolumn{6}{c}{CHAIR$_\text{i}(\downarrow)$} & \multirow{2}{*}{Average length} \\
          \cmidrule(lr){3-8}\cmidrule(lr){9-14}
          &  & $32$ & $64$ & $128$ & $256$ & $512$ & $1024$ & $32$ & $64$ & $128$ & $256$ & $512$ & $1024$ & \\
          \midrule
         \multirow{8}*{\rotatebox[origin=c]{90}{Llava \cite{llava1}}}& Original & $7.8$ & $23.8$ & $50.0$ & $50.6$ & $50.6$ & $50.6$ & $4.5$ & $8.2$ & $15.3$ & $15.3$ & $15.3$ & $15.3$ & $500$ \\
         & \textbf{+ReCo} & \textcolor{NEGATIVE}{+$4.6$} & \textcolor{NEGATIVE}{+$0.4$} & \textcolor{POSITIVE}{-$1.4$} & \textcolor{POSITIVE}{-$5.4$} & \textcolor{POSITIVE}{-$16.8$} &
         \textcolor{POSITIVE}{-$22.6$} & \textcolor{NEGATIVE}{+$2.8$} & \textcolor{NEGATIVE}{+$1.3$} & \textcolor{POSITIVE}{-$0.3$} & \textcolor{POSITIVE}{-$0.2$} &
         \textcolor{POSITIVE}{-$3.9$} & \textcolor{POSITIVE}{-$5.1$}  & $319$ \\
         \cdottedline{2-15}
         % \cdashline{2-15}[2pt/3pt]\\[-2ex]
         & VCD \cite{vcd} & $8.0$ & $22.6$ & $52.2$ & $48.4$ & $48.4$ & $48.5$ & $4.4$ & $7.7$ & $16.0$ & $15.3$ & $15.3$ & $15.3$ & $486$  \\
         & \textbf{+ReCo} & \textcolor{POSITIVE}{-$0.4$} & \textcolor{NEGATIVE}{+$2.4$} & \textcolor{POSITIVE}{-$10.4$} & \textcolor{POSITIVE}{-$8.0$} & \textcolor{POSITIVE}{-$15.9$} & \textcolor{POSITIVE}{-$13.6$} & \textcolor{POSITIVE}{-$0.7$} & \textcolor{NEGATIVE}{+$1.1$} & \textcolor{POSITIVE}{-$3.1$} & \textcolor{POSITIVE}{-$2.9$} & \textcolor{POSITIVE}{-$3.3$} & \textcolor{POSITIVE}{-$4.0$}  & $306$ \\
         \cdottedline{2-15}
         % \cdashline{2-15}[2pt/3pt]\\[-2ex]
         & M3ID \cite{m3id} & $8.2$ & $22.4$ & $55.4$ & $57.2$ & $57.2$ & $57.2$ & $3.9$ & $6.9$ & $15.8$ & $16.2$ & $16.2$ & $16.2$  & $495$  \\
         & \textbf{+ReCo} & \textcolor{NEGATIVE}{+$1.2$} & \textcolor{POSITIVE}{-$0.2$} & \textcolor{POSITIVE}{-$6.2$} & \textcolor{POSITIVE}{-$16.8$} & \textcolor{POSITIVE}{-$19.0$} & \textcolor{POSITIVE}{-$24.8$} & \textcolor{NEGATIVE}{+$0.7$} & \textcolor{NEGATIVE}{+$0.7$} & \textcolor{POSITIVE}{-$2.6$} & \textcolor{POSITIVE}{-$4.2$} & \textcolor{POSITIVE}{-$5.2$} & \textcolor{POSITIVE}{-$5.7$} & $335$\\
         \cdottedline{2-15}
         % \cdashline{2-15}[2pt/3pt]\\[-2ex]
         & AvisC \cite{avisc} & $9.2$ & $21.0$ & $53.6$ & $55.8$ & $55.8$ & $55.8$ & $5.3$ & $6.9$ & $15.3$ & $17.1$ & $17.1$  & $17.1$ & $523$ \\
         & \textbf{+ReCo} & \textcolor{POSITIVE}{-$0.1$} & \textcolor{NEGATIVE}{+$2.0$} & \textcolor{POSITIVE}{-$0.2$} & \textcolor{POSITIVE}{-$8.8$} & \textcolor{POSITIVE}{-$9.2$} & \textcolor{POSITIVE}{-$9.8$} & \textcolor{POSITIVE}{-$0.3$} & \textcolor{NEGATIVE}{+$1.1$} & \textcolor{NEGATIVE}{+$1.3$} & \textcolor{POSITIVE}{-$2.7$} & \textcolor{POSITIVE}{-$3.6$} & \textcolor{POSITIVE}{-$3.5$} & $428$ \\
         \bottomrule
    \end{NiceTabular}
    
\end{table*}

\setlength{\tabcolsep}{6pt} % Default value: 6pt

\subsection{MME}

Besides POPE, which is restricted to existence-related questions, we evaluate ReCo in the context of multiple types of discriminative questions, using the MME benchmark. Although, just like POPE, such a benchmark does not directly assess the effect (\textit{fading memory effect}) we are trying to eradicate in this work, it is important to examine whether the addition of ReCo (or any other component) hurts the model's performance in such discriminative tasks. In \cref{tab:mme}, we can observe that ReCo not only preserves but rather improves the performance of InstructBLIP \cite{instructblip} and MiniGPT4 \cite{minigpt4}, with or without the addition of any of the other baselines. Regarding Llava \cite{llava1}, the addition of ReCo preserves the already amazing performance, while at the same time, as we showed in \cref{tab:chair_llava}, it reduces dramatically its hallucinating performance on the free text generation.

\begin{table}[h!]
    \scriptsize
    \centering
    \caption{MME \cite{amber} results for InstructBLIP \cite{instructblip}, MiniGPT4 \cite{minigpt4}, and Llava \cite{llava1}. ReCo consistently improves the performance of InstructBLIP and MiniGPT4, while the performance of Llava remains at the same level.}
    \label{tab:mme}
    \begin{NiceTabular}{l c c c }
         \toprule 
         & \multicolumn{3}{c}{MME ($\uparrow$)} \\
         \cmidrule(lr){2-4}
         Model (\textcolor{POSITIVE}{+ReCo}) & InstructBLIP & MiniGPT4 & LLaVA  \\
        \midrule
         Original & $1355$ (\textcolor{POSITIVE}{+$150$}) & $939$ (\textcolor{POSITIVE}{+$051$}) & $1543$ (\textcolor{NEGATIVE}{-$019$})  \\
         % \cdashline{1-3}[2pt/3pt]\\[-2ex]
         VCD \cite{vcd} & $1495$ (\textcolor{POSITIVE}{+$062$}) & $933$ (\textcolor{POSITIVE}{+$115$})  & $1582$ (\textcolor{POSITIVE}{+$045$})  \\
         % \cdashline{1-3}[2pt/3pt]\\[-2ex]
         M3ID \cite{m3id}  & $1458$ (\textcolor{POSITIVE}{+$080$}) & $961$ (\textcolor{POSITIVE}{+$061$})  & $1619$ (\textcolor{POSITIVE}{+$031$})  \\
        % \textbf{+ReCo} & \textcolor{POSITIVE}{+$3.5$} & \textcolor{POSITIVE}{+$33.1$}\\
        %  \cdashline{1-3}[2pt/3pt]\\[-2ex]
         AvisC \cite{avisc}  & $1373$ (\textcolor{POSITIVE}{+$121$}) & $879$ (\textcolor{NEGATIVE}{-$023$})  & $1661$ (\textcolor{NEGATIVE}{-$051$})  \\
        % \textbf{+ReCo} & \textcolor{POSITIVE}{+$2.2$} & \textcolor{NEGATIVE}{-$4.6$}\\
         \bottomrule
    \end{NiceTabular}
    
\end{table}

\section{More qualitative results}

In the following figures, we present more examples for both generative and discriminative questions. In many of the cases, the VLMs fail to provide a cohesive answer. On the contrary, ReCo is able to significantly improve the results, as it is apparent by the reduction of hallucinating objects, unrelated-to-image text, as well as by the accurate answering of the discriminative questions.

\begin{figure*}
    \centering
    \includegraphics[width=\textwidth]{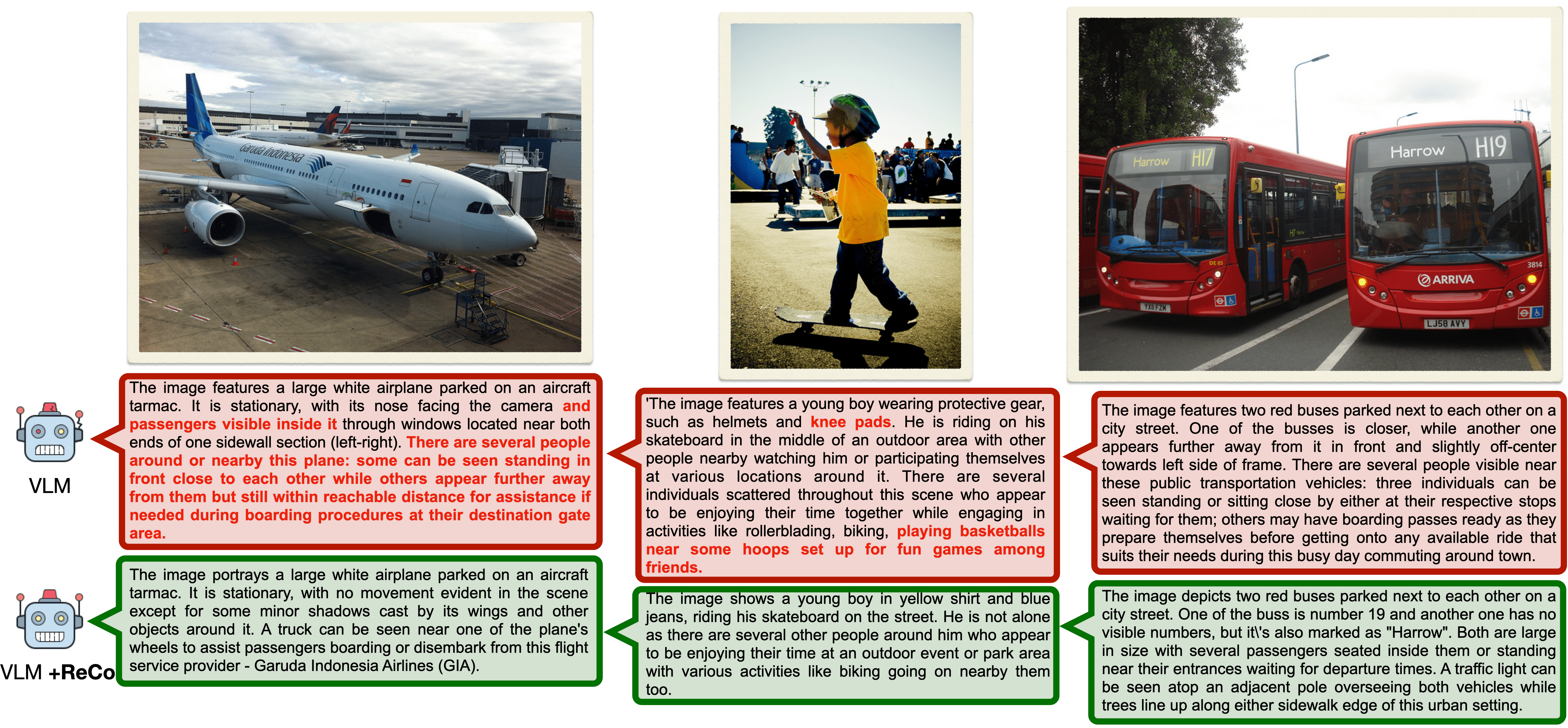}\hfill 
    \includegraphics[width=\textwidth]{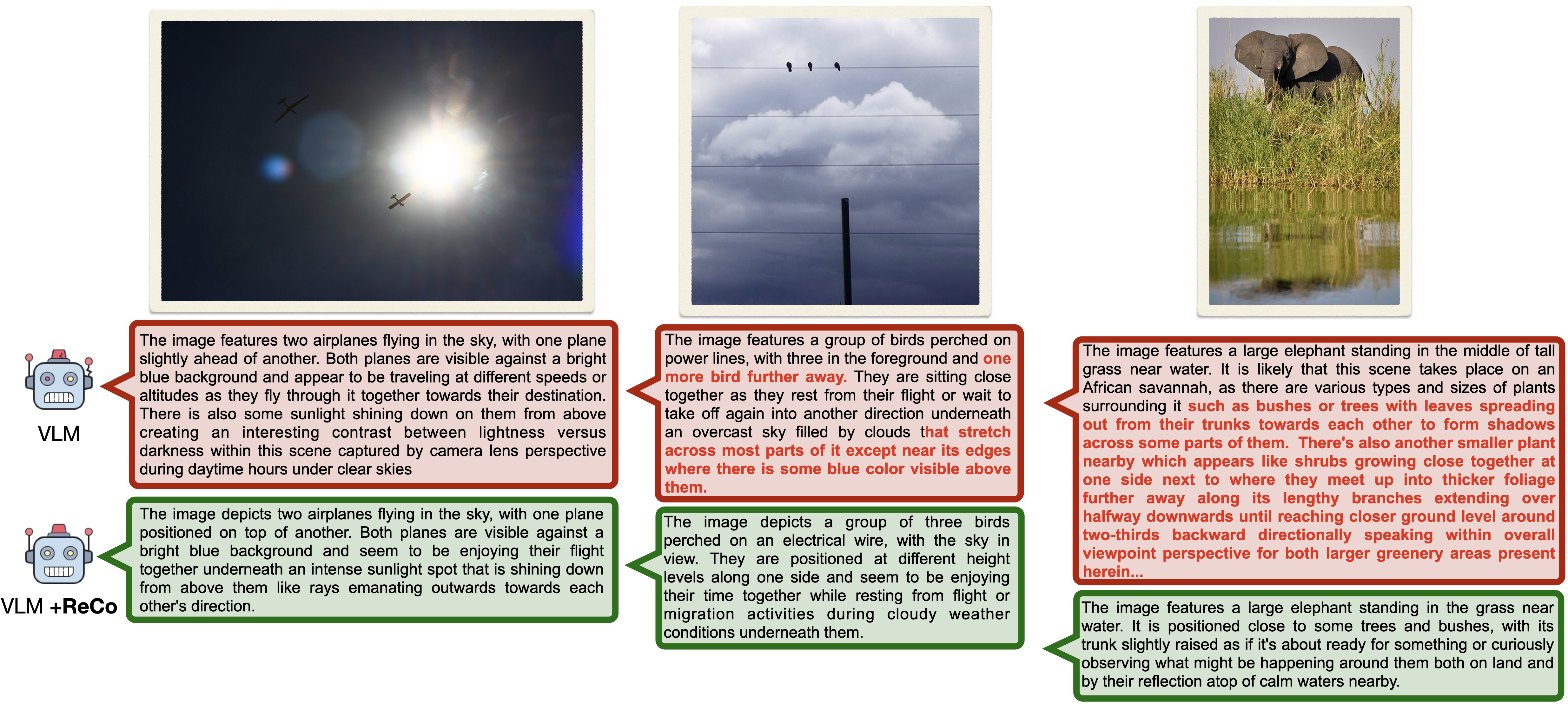}
    \caption{Generative questions on InstructBLIP \cite{instructblip}. The prompt used is \textit{``Describe the image.''}}
    \label{fig:enter-label}
\end{figure*}

\begin{figure*}
    \centering
    \includegraphics[width=\textwidth]{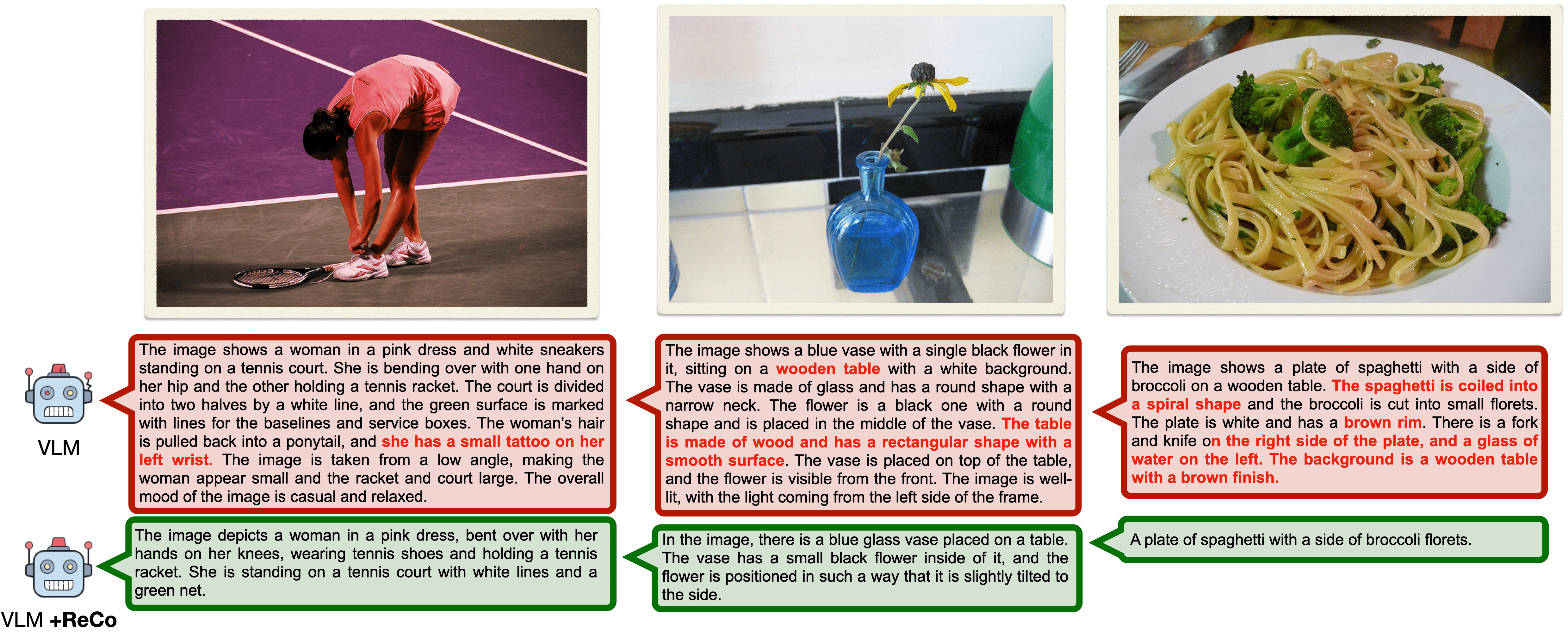}\hfill 
    \includegraphics[width=\textwidth]{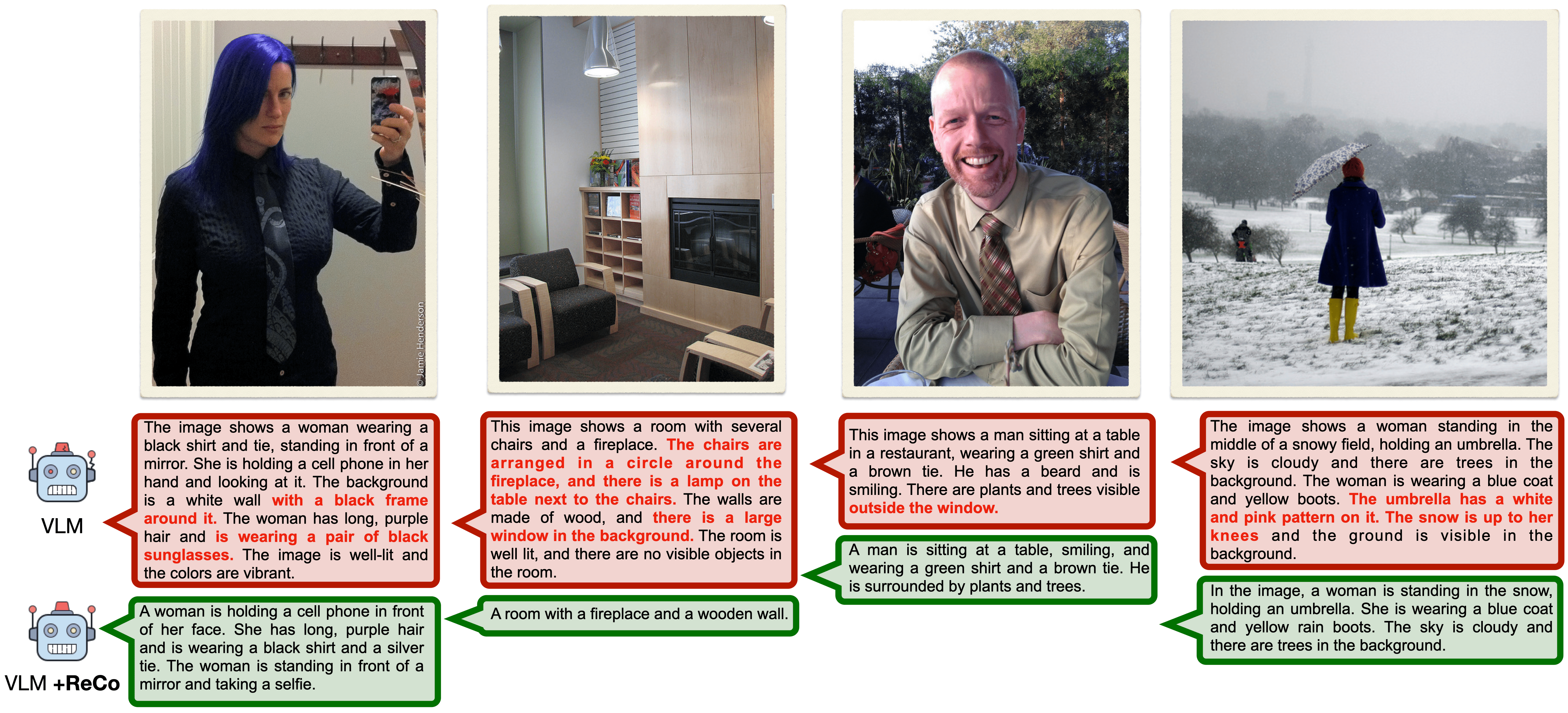}
    \caption{Generative questions on MiniGPT4 \cite{minigpt4}. The prompt used is \textit{``Describe the image.''}}
    \label{fig:enter-label}
\end{figure*}

\begin{figure*}
    \centering
    \includegraphics[width=\textwidth]{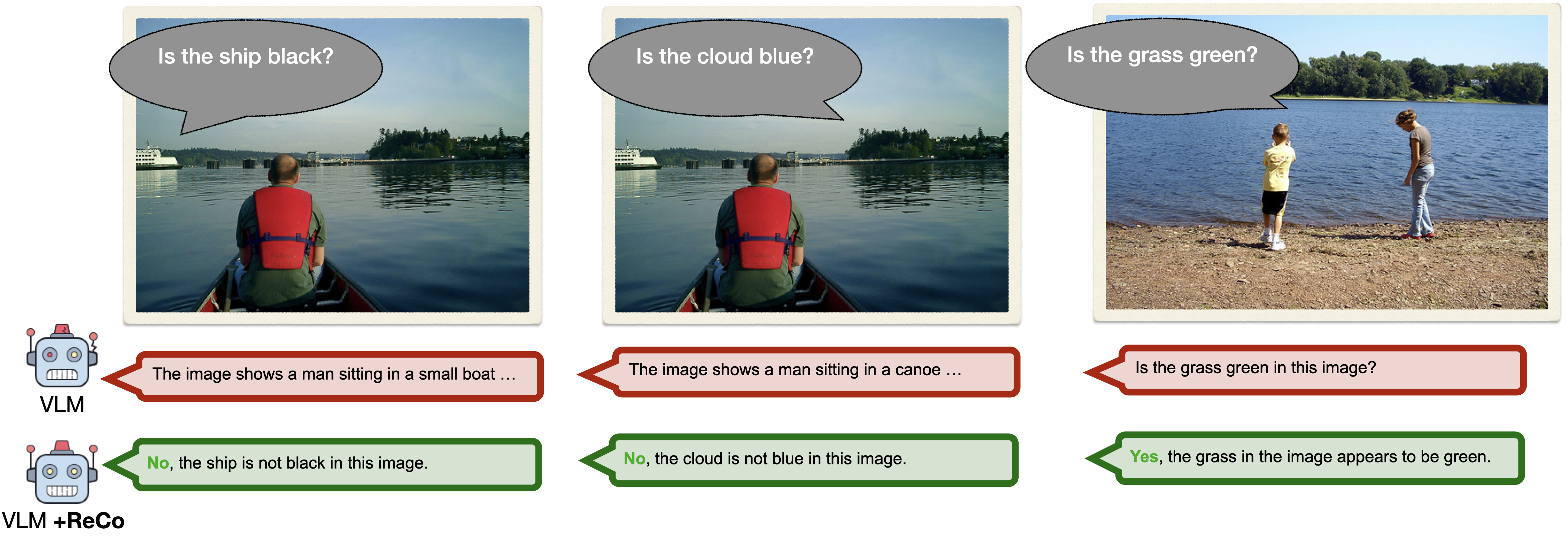}\hfill 
    \includegraphics[width=\textwidth]{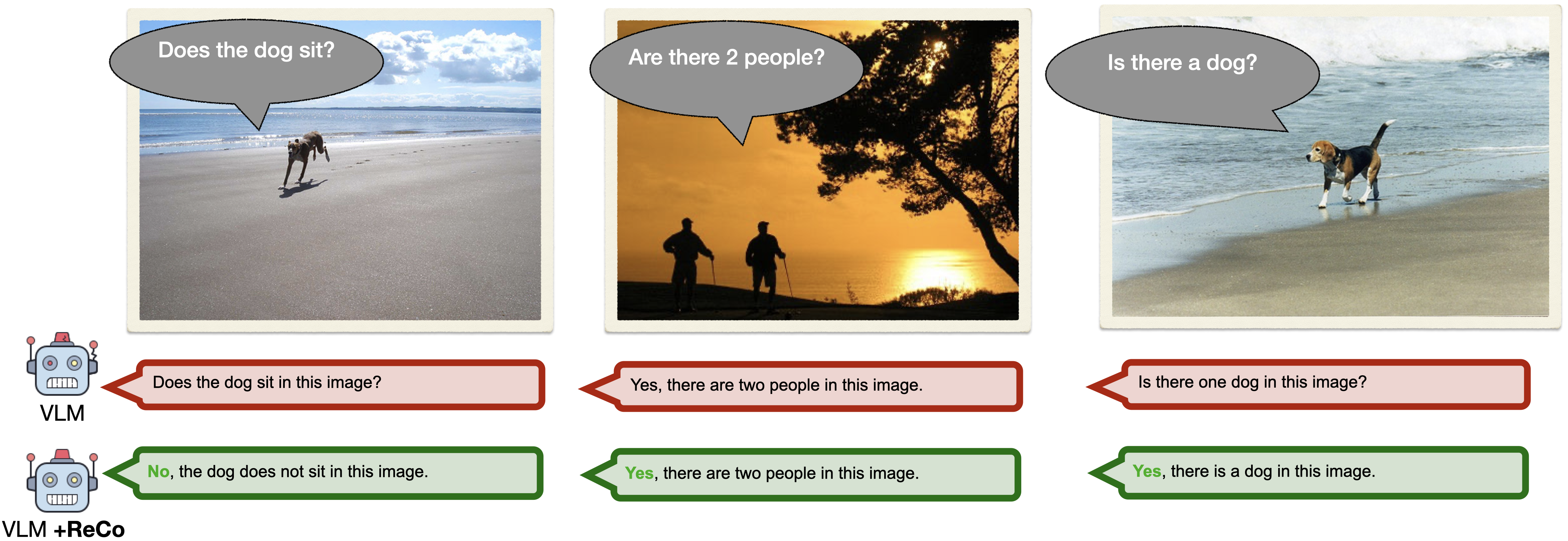}
    \includegraphics[width=\textwidth]{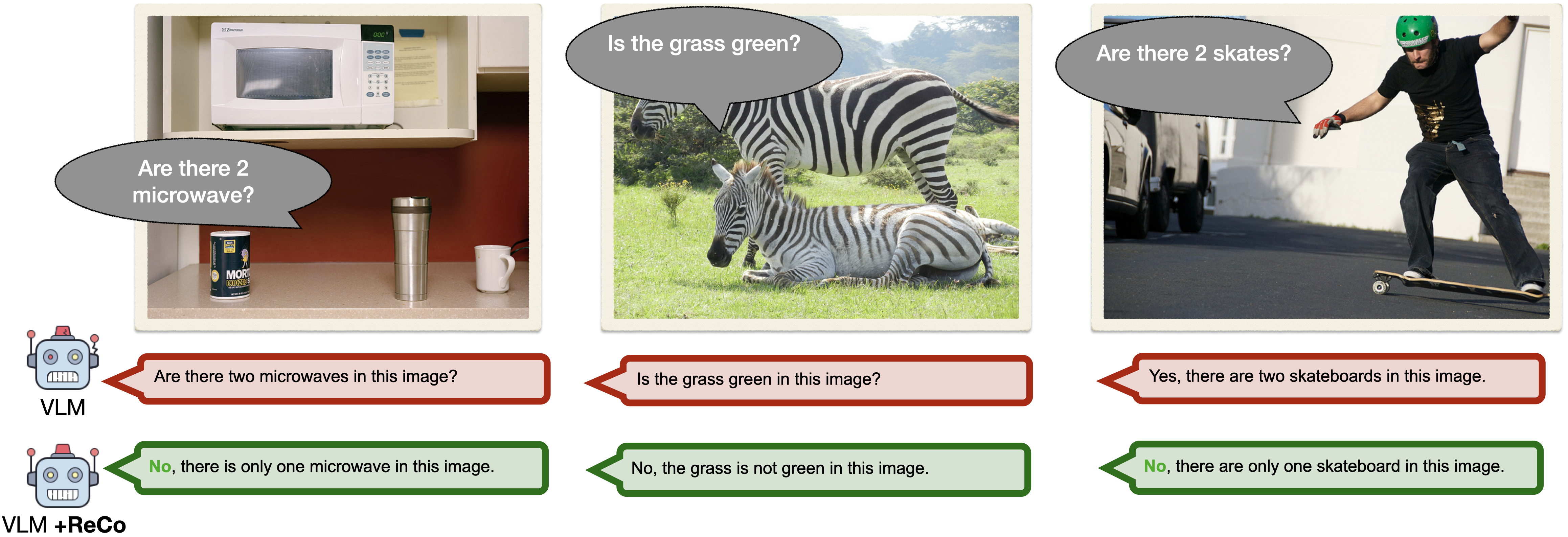}
    \caption{Discriminative questions on MiniGPT4 \cite{minigpt4}, taken from the AMBER benchmarks. The unmodified VLM fails to answer cohesively in most of the cases, while the addition of ReCo (although not trained on such data) allows the model to answer correctly.}
    \label{fig:enter-label}
\end{figure*}

\end{document}